\documentclass[fleqn, 10pt]{wlscirep}
\usepackage{wasysym}
\usepackage{booktabs}
\usepackage{graphicx}
\usepackage{multirow}
\usepackage{subcaption}
\usepackage{nicefrac}
\usepackage{subfiles}

\newcommand{\etal}{\textit{et al}. }
\newcommand{\ie}{\textit{i}.\textit{e}., }
\newcommand{\eg}{\textit{e}.\textit{g}. }

\newcommand\doublecheck{{\checked\kern-0.6mm\checked}}

\title{Gaze-in-wild: A dataset for studying eye and head coordination in everyday activities}
\author[1*]{Rakshit Kothari}
\author[2]{Zhizhuo Yang}
\author[1$\dagger$]{Christopher Kanan}
\author[2$\dagger$]{Reynold Bailey}
\author[1$\dagger$]{Jeff Pelz}
\author[1$\dagger$]{Gabriel Diaz}
\affil[1]{Chester F. Carlson Center for Imaging Science, RIT}
\affil[2]{Golisano College of Computing and Information Sciences, RIT}
\affil[*]{\textbf{email:} rsk3900@rit.edu}
\affil[$\dagger$]{These authors contributed equally to this work}
\makeindex

\begin{abstract}

The interaction between the vestibular and ocular system has primarily been studied in controlled environments. Consequently, off-the shelf tools for categorization of gaze events (\eg fixations, pursuits, saccade) fail when head movements are allowed. Our approach was to collect a novel, naturalistic, and multimodal dataset of eye+head movements when subjects performed everyday tasks while wearing a mobile eye tracker equipped with an inertial measurement unit and a 3D stereo camera. This Gaze-in-the-Wild dataset (GW) includes eye+head rotational velocities (deg/s), infrared eye images and scene imagery (RGB+D). A portion was labelled by coders into gaze motion events with a mutual agreement of 0.72 sample based Cohen's $\kappa$. This labelled data was used to train and evaluate two machine learning algorithms, Random Forest and a Recurrent Neural Network model, for gaze event classification. Assessment involved the application of established and novel event based performance metrics. Classifiers achieve $\sim$90$\%$ human performance in detecting fixations and saccades but fall short (60$\%$) on detecting pursuit movements. Moreover, pursuit classification is far worse in the absence of head movement information. A subsequent analysis of feature significance in our best-performing model revealed a reliance upon absolute eye and head velocity, indicating that classification does not require spatial alignment of the head and eye tracking coordinate systems. The GW dataset, trained classifiers and evaluation metrics will be made publicly available with the intention of facilitating growth in the emerging area of head-free gaze event classification.

\end{abstract}

\begin{document}
\maketitle

\section{Introduction}


Human visual behavior can be viewed as a sequence of periods of stable visual input, punctuated by saccades to new locations within the visual environment.  Although saccadic targeting during visual search may demonstrate an influence of visual salience,~\cite{Itti1998AAnalysis} the effect is overwhelmed in the presence of a task, when motor execution requires that attention be directed towards information-rich, task-relevant locations within the visual environment.~\cite{Tatler2010YarbusVision, Hayhoe2012,Hayhoe2005EyeBehavior} As a result, the dynamics of gaze coordination in natural contexts are affected by a variety of extra-retinal properties of the task, the agent, the environment, and by their interaction.  These include the spatial distribution of information in the natural environment,~\cite{Sprague2015StereopsisEnvironment} cognitive resources related to memory or higher order reasoning,~\cite{Kothari2016NovelObstacles} motor constraints that determine the dynamics of gaze shifts,~\cite{Daye2014 , Barnes1979Vestibulo-ocularTargets., Freedman2008CoordinationOrienting,Einhauser2007} and biomechanical constraints that influence visual strategies for foot placement during locomotion.~\cite{Matthis2018GazeTerrain}

Despite the importance of extra-retinal influences upon gaze behavior during visually guided action, surprisingly little attention has been dedicated to the study of gaze behavior in more natural contexts. For instance, head movements are often suppressed through the use of a chin-rest, or by constraining target movement to only a small portion of the subject's visual field. Furthermore, target motion is often restricted to two dimensions, and sometimes viewed monocularly. In part, the study of strategies for coordination of the eyes, head, and body has been limited by a lack of suitable technology. Successful tracking of the coordination between the head and eyes in unconstrained settings requires advances in two parallel domains: the instrumentation to jointly monitor the direction of the eyes and head (``eye + head tracking''), and the algorithms to parse and categorize key oculomotor events in the rapid stream of data (i.e. ``event detectors'').

The present study aims to develop new event detectors for the study of eye and head coordination during natural behavior. This involves both the development of a custom eye+head tracker, and the capture of a novel dataset of head-free gaze behavior - the Gaze-In-Wild dataset (GW). GW was collected from 19 participants engaged in everyday activities using spatially and temporally calibrated equipment comprised of a hardhat with an inertial measurement unit (IMU), eye tracking glasses, and a stereo-based RGB-D (RGB imagery plus depth) sensor. A custom-made software tool which facilitates the efficient hand-labelling of the captured data was used to label a significant portion (approx. 2 hours and 15 minutes) of the GW dataset (see Section~\ref{Labelling}). We then use this labelled data for supervised training and assessment of automated event detectors.


This work builds upon a variety of techniques previously used to track head orientation during natural behavior. Published studies have demonstrated the use of rotational potentiometers and accelerometers,~\cite{Barnes1979Vestibulo-ocularTargets.} magnetic coils,~\cite{Epelboim1995TheTasks} or motion capture~\cite{fang2015eye} for the sensing of head orientation.~\cite{Kothari2016NovelObstacles} Perhaps the highest precision eye+head tracker which allowed body movement leveraged a 5.8 $m^3$ custom-made armature capable of generating a pulsing magnetic field.  The subject was outfitted with a head-worn receiver capable of measuring head position and orientation within its operational region.~\cite{Allison1996CombinedSystem} Several systems have adopted video based head motion compensation~\cite{Kinsman2012Ego-motionTracking, Einhauser2007} and demonstrated promising results, but are too computationally expensive for real-time use, and are prone to irrecoverable track loss especially during periods of rapid head movement, occlusion of tracking features or degration of image quality due to motion blur. Recent approaches have involved the use of head-mounted IMUs. For example, Larsson \etal used a head-mounted IMU in a study where subjects were asked to perform visual tracking tasks when watching pre-rendered stimuli projected onto a 2D screen.~\cite{Larsson2016} They established that compensating for head movements results in a reduced standard deviation for the eye position signal. More recently, Tomasi \etal used two IMUs for tracking eye and head orientation relative to heading direction~\cite{Tomasi2016MobileStudies} and reported a 7.1$^\circ$ average angular error ($\sigma$=5.2$^\circ$).

This work also builds upon a long history of methodologies for the automated detection of gaze events within an eye tracking signal. The simplest methods use threshold based filters and numerous descriptive features for classification.~\cite{holmqvist2011eye} Threshold based techniques require parameter tuning for each test scenario as well as being sensitive to noise and sample rate. A better solution is to use machine learning to learn a model for classifying gaze events. These algorithms have been shown to work well when the head is fixed. Pekkanen \etal proposed the Naive Segmented Linear Regression (NSLR) model~\cite{Pekkanen2017ARegression} which segments a time sequence into distinguishable events which are then classified using continuous Hidden Markov Models (HMM). While earlier work used hand-crafted features,~\cite{Zemblys2018UsingData} more recent methods have employed recurrent neural networks (RNN)~\cite{Zemblys2018GazeNetNetworks} which enable algorithms to directly learn what features are relevant to the task.  However, both approaches have traditionally only been designed for situations when the head is fixed. New gaze classification algorithms are needed for datasets that incorporate both head and eye movements.

\subsection{Classification scheme and nomenclature}
\label{label_scheme}

\begin{figure}
    \centering
    \includegraphics[width=\textwidth]{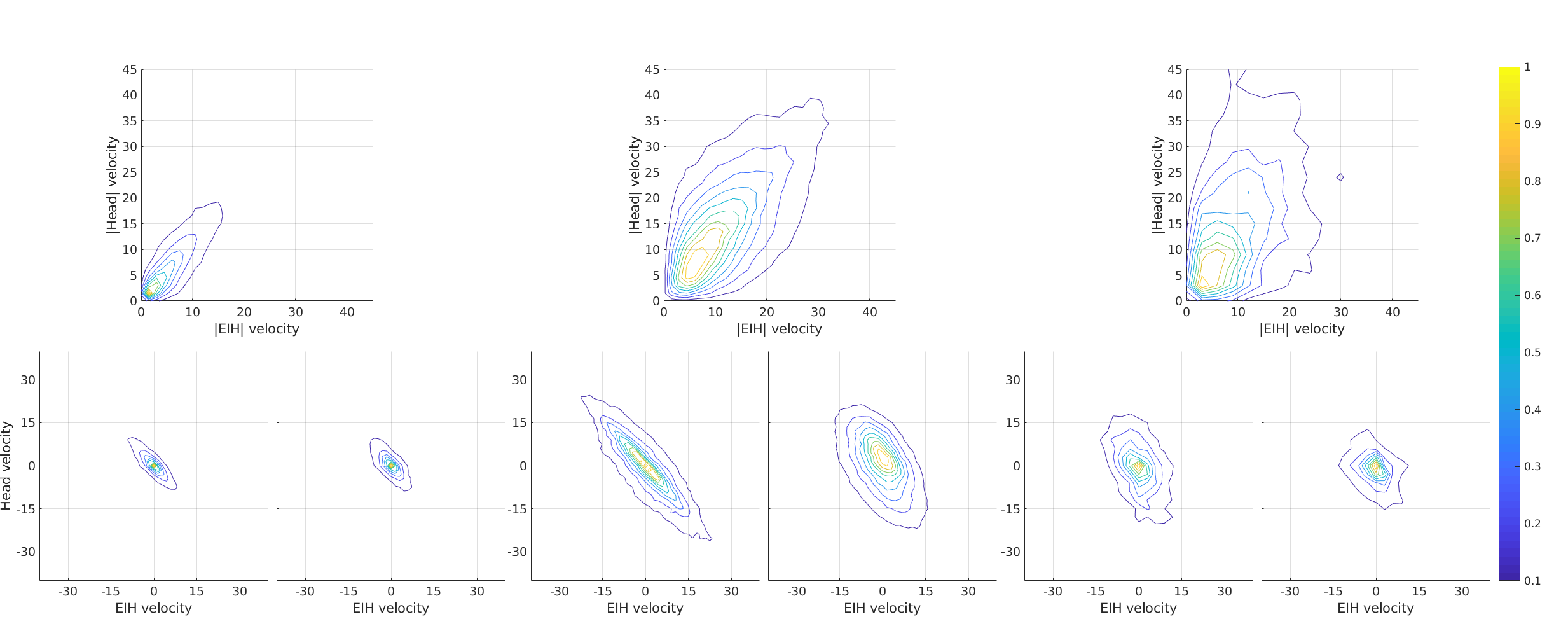}
    \caption{Eye and head movement statistics. The top row signifies absolute eye and head velocity. The bottom row signifies the distribution of eye and head velocity in the azimuth and elevation direction. The  left column illustrate fixations when subjects were stationary. The middle column illustrates fixations when subjects were in translatory motion. The right column illustrates pursuit behavior. The scale on the right shows the normalized concentration of samples and is used in all figures.}
    \label{fig:Eye&Head_stats}
\end{figure}

Gaze classification requires distinct and separable classes that are identifiable in our daily activities. There has been some disagreement in the research community about the specific criteria for establishing a taxonomy of gaze events.\cite{Lappi2016EyeReference, Nystrom2018IsResearchers} For example, one approach is to classify events based upon specific oculomotor movements, such as the two major retinal image stabilizing mechanisms: the vestibular-ocular response (VOR), and the opto-kinetic response (OKR). In VOR, the semicircular canals of the inner ear measure head rotation acceleration which results in eye movements in the opposite direction with near unity gain (\ie the ratio of eye and head velocity is $\sim$1, see Figure~\ref{fig:Eye&Head_stats}). OKR is generated by retinal motion which in turn leads to compensatory eye movements to reduce retinal blur.~\cite{Land2009TheRepertoire, Barnes1989Head-freeManner., Barnes1979Vestibulo-ocularTargets., Barnes1993Visual-vestibularMechanisms}. It is difficult to derive a classification scheme based solely on these stabilizing mechanisms, because they may be used in isolation, or in combination, for either fixation of a target that is stationary in the exocentric frame, or pursuit of a moving target. 

Our approach is to adopt an exocentric classification scheme and to discuss its applicability in classifying a broad range of coordinated head and eye movements. We define movement categories by the functional role of the eye movement, as well as the motion of an object within an exocentric frame of reference. As a result, events in our dataset is classified as follows: 


\begin{enumerate}
\item \textbf{Gaze fixation (GF)} - Gaze fixation may be brought about through stabilization of the eyes and head, or during movements of the eyes and head that are compensatory and, as a result, produce a stable gaze vector on a stationary object in the world coordinate frame. Stabilized retinal image motion lies near to the range of 0.5 to 5 $^\circ/s$, a limit above which the target image starts to blur.~\cite{Land2009TheRepertoire} Hence, a wide range of miniature head compensated eye movements can be termed as gaze fixation. In our taxonomy, gaze fixations may be further categorized as:
\begin{itemize}
    \item \textbf{Tremors} - The resting eye and head rarely display perfect stability. Skavenski \etal identified that despite instructing subjects to remain as stationary as possible, tremor was observed in the head and eyes ($<$1$^\circ/s$, 10$Hz$).~\cite{A.A.Skavenski1979QualityMan} Furthermore, the characteristics of tremor is known to vary based on the nature of the instrumentation~\cite{Martinez-Conde2004ThePerception} and type of restraint.~\cite{A.A.Skavenski1979QualityMan}
    
    \item \textbf{Drift} - Drifts are slow motions of the eye that are often punctuated by microsaccades and aid in maintaining crisp visual features across the retina. While there is some disagreement on the range of drift motion, they usually display amplitudes within 0.25$^\circ$ and velocities less than 0.5$^\circ/s$ when the head is fixed.~\cite{Martinez-Conde2004ThePerception}
   
    \item \textbf{Microsaccades} - Small, rapid eye movements that occur in between fixations are termed as microsaccades and usually last about 25$ms$ with a velocity range capped at 50$^\circ/s$.~\cite{Martinez-Conde2004ThePerception}
    
    \item \textbf{Fixation by rotational vestibular-ocular reflex (rVOR)} - When the subject and target are stationary in the world reference frame, rotational motion of the head is compensated using rVOR. Fixations are maintained by the VOR system because it has a significantly lower response lag as compared to OKR.~\cite{Land2009TheRepertoire} Generally, a rVOR event displays near unity gain unless it is modulated due to other compensatory mechanisms such as OKR or pursuit.
    
    \item \textbf{Fixation by translational vestibular-ocular reflex (tVOR)} - When a target is stationary in the world reference frame, image stability at the fovea during self motion or passive displacements is achieved by tVOR~\cite{Angelaki2004EyesMotion}. Unlike rVOR, wherein a counter rotation of the eye in head rotation can stabilize the entire retinal image, tVOR cannot accommodate for the entire visual field due to the large range of optic-flow motion experienced at different depth planes. Primarily a foveal image adjustment mechanism, it follows that properties of tVOR motion depend on the gaze direction and can be difficult to differentiate with pursuit movements.~\cite{Angelaki2006Three-DimensionalConstraints} During low target retinal velocity ($\sim$0$^\circ/s$, $<$0.5$Hz$),~\cite{Land2009TheRepertoire} the vestibulo-ocular reflex gain is not adequate to achieve image stability.~\cite{Barnes1993Visual-vestibularMechanisms} OKR augments VOR to help maintain a stable image over stationary targets. Fixations are maintained by a combination of gain modulation and optokinetic stimulation.~\cite{Barnes1993Visual-vestibularMechanisms, Land2009TheRepertoire, Angelaki2004EyesMotion, mustari2010optokinetic} While microsaccades may be triggered for retinal image adjustment, larger saccades during fixations signify shifts in attention or an inability of gain adjustment to compensate for motion such as observed during nystagmus. These visually driven eye movements work in synergy with tVOR~\cite{Angelaki2006Three-DimensionalConstraints} making them difficult to observe in everyday activities as opposed to controlled experiments which are designed to isolate their behavior.
\end{itemize}

\item \textbf{Gaze pursuit (GP)} - Also known as smooth pursuit movements,~\cite{Ackerley2011} gaze pursuit is the visual tracking of an object that is moving through the world frame using the eyes or a combination of the eyes and head by augmenting over our compensatory systems.~\cite{Land2009TheRepertoire} Gaze pursuit is often interrupted by catch-up saccades in compensation of retinal error.~\cite{Daye2014} While it is somewhat trivial to identify GP events using visual imagery, it may become difficult to differentiate them with GF (for more information refer to Supplementary Figure~\ref{fig:example_signals}).

\item \textbf{Gaze shift (S)} - A rapid shift of the fovea to a new location in the world brought about by the eye alone (i.e. a saccade).
\end{enumerate}

To illustrate our nomenclature, consider a situation where a person under fore-aft motion attempts to pursue a moving target. In situations such as these, the effects of stepping are compensated using VOR in the elevation direction. Relative distance and gaze angle modulates the tVOR to maintain target image at the fovea. The moving target's retinal image motion elicits a pursuit signal punctuated by predicative saccades. The pursuit motion augments over translational VOR by modulating its gain. If the eye and head pursuit movement can be distinctly identified in their velocity traces, we would consider such an event as a gaze pursuit. However, a distant or slow moving target may induce a small pursuit signal which may not be easily identifiable over opto-kinetic stabilization of the retinal image. In these situations, we would consider the event as a gaze fixation.

\section{The Gaze-in-wild dataset}
\label{Methods}
\begin{figure}
    \centering
    \includegraphics[width=0.75\textwidth]{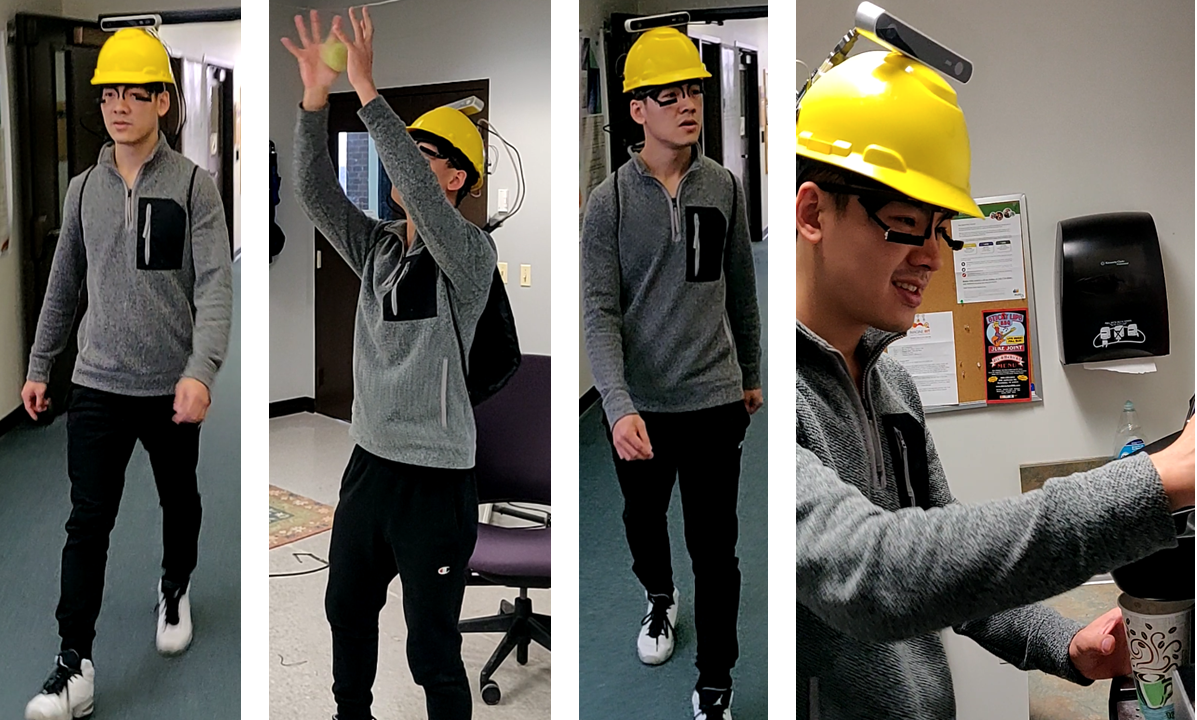}
    \caption{Task selections in the GW dataset. Left to right $\rightarrow$ Indoor navigation, ball catching, visual search and tea making.}
    \label{fig:setup_with_tasks}
\end{figure}

\begin{figure}
    \centering
    \includegraphics[width=1\linewidth]{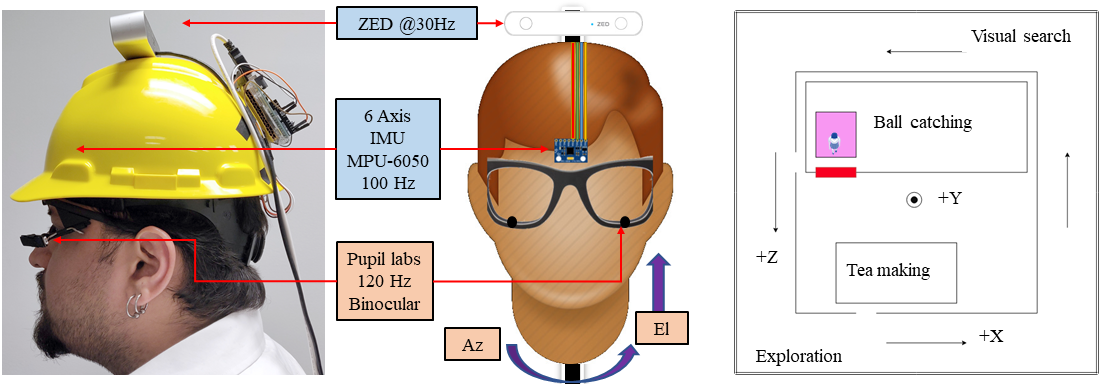}
    \caption{(left) side-view, (middle) front view of hardware setup. (right) Top view of all trajectories within our world coordinate system. The red box indicates the position of the calibration pattern. The purple box signifies the region where subject stood during calibration.}
    \label{fig:HW_setup}
\end{figure}

\begin{figure}
  \includegraphics[width=\linewidth]{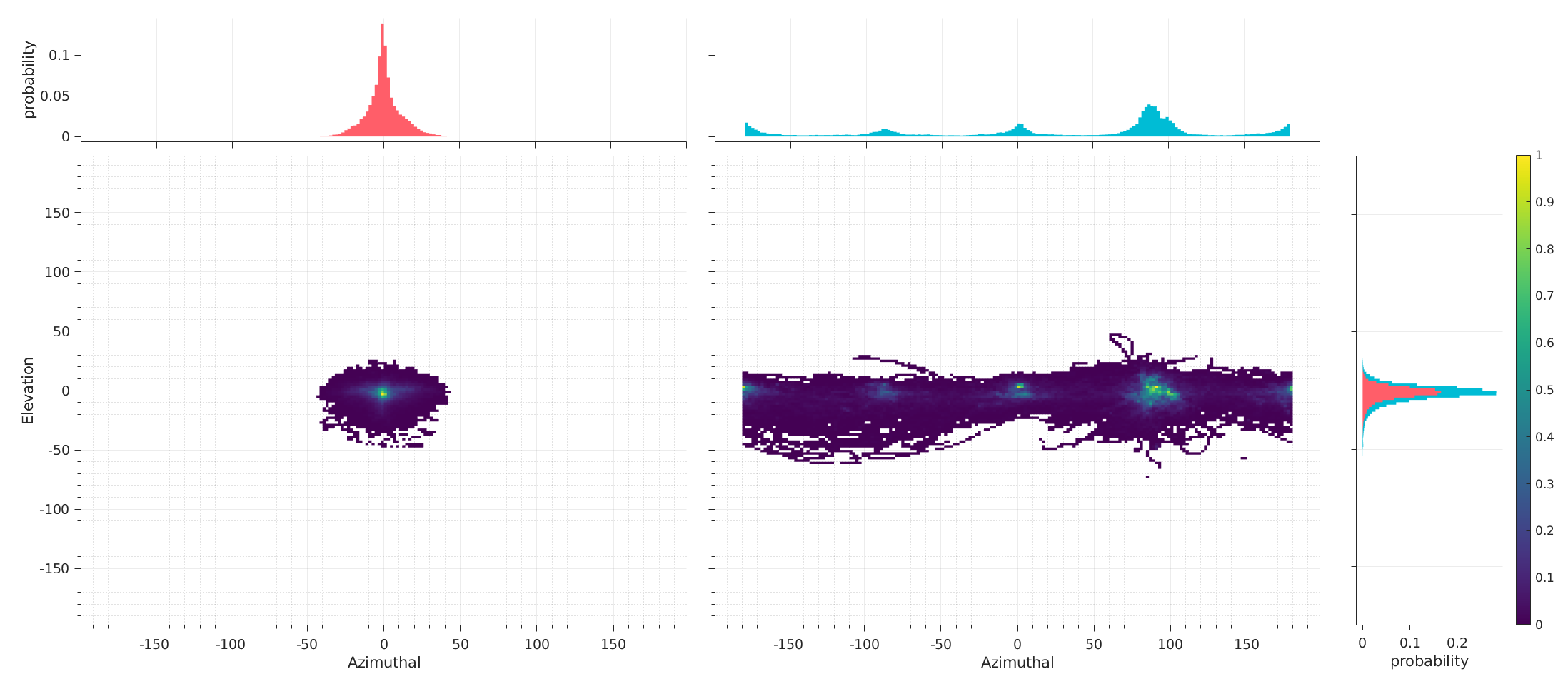}
  \caption{Head pose and cyclopean eye distribution in the azimuthal and elevation direction. Angles are provided in degrees. Note that head distribution peaks occur at 90$^\circ$ intervals.}
  \label{fig:head&eye_dist}
\end{figure}

The aim of this work is to generate a dataset that captures complex ocular-motor strategies during natural tasks (see Figure~\ref{fig:setup_with_tasks}). We recruited 19 participants (7 female, age $\mu$=28, $\sigma$=12.52). Informed consent was obtained from all participants prior to hardware setup to anonymously share all data recorded from them. All methods were carried out in accordance with relevant guidelines and regulations as approved by the Institutional Review Board at Rochester Institute of Technology, FWA-00000731. Participants were tasked with performing up to four activities while wearing an eye tracker, a hardhat instrumented with sensors, and a backpack with a laptop computer (see Figure~\ref{fig:HW_setup}). Since task demands and interpretation have been known to guide eye movements,~\cite{Tatler2010YarbusVision} care was taken to ensure all participants received a standard set of instruction read aloud by the experimenter. Subjects were instructed to stand 1 to 2 meters away from a calibration chart within a predefined rectangular area. Once a participant was within the calibration region and facing the chart, they performed two calibration routines. After calibration was complete, participants proceeded to complete the given task. Table~\ref{tbl:Dataset_status} in the Supplementary lists the calibration accuracy, tasks recorded, and the labelling status of each observer. Tasks were selected to create a wide range of head and eye poses as seen in Figure~\ref{fig:head&eye_dist}. Upon completion of a task, participants returned to the calibration area to prepare for the next task. The following tasks were chosen:
\begin{itemize}
    \item \textbf{Indoor navigation}:
    Subjects were instructed to walk around an indoor corridor loop twice. Indoor navigation was chosen to elicit coordinated eye and head movements that occur naturally during walking. We observed various gaze shifts to objects such as text on posters, signboards, people walking by etc. As expected, subjects made very few to no gaze shifts towards the ground due to lack of terrain complexity \cite{Matthis2018GazeTerrain} and very little attention demands \cite{Kothari2016NovelObstacles} for foot placement accuracy \cite{Matthis2014VisualTerrain}. While some of the subjects were familiar with the indoor corridor layout, we did not observe any noticeable difference in their behavior compared to subjects unfamiliar with the environment.
    \item \textbf{Ball catching}:
    The purpose of this task was to induce gaze pursuit behavior by asking participants to play catch with the experimenter. The experimenter would change throwing strategies in the middle of the task by either bouncing the ball on the floor, passing the ball to another experimenter or rolling the ball on the ground towards the participant. The subjects tracked the ball as a series of gaze fixations and predictive catch-up/look-ahead saccades and occasionally pursued the ball during a specific period of the ball trajectory.
    \item \textbf{Object search without prior subject-object interaction}: Subjects were tasked to locate and count as many objects with geometrical shapes (such as triangles, rectangles etc.) as they could find in a predetermined closed circuit corridor. This task was chosen to elicit visual search behavior in a head-free setting without biasing a subject with a particular object or shape.
    \item \textbf{Tea making}: As a validation for the classic tea making paradigm,~\cite{Land2001InActivities} we instructed subjects to go to the kitchen and make themselves a cup of tea. For this task, due to the close proximity of objects, relevant information sometimes fell outside the field of view.
\end{itemize}

\subsection{Hardware setup and error categorization}
\label{Methods:HW}
To collect naturalistic data, we instrumented participants with an MPU-6050 6-axis Inertial Measurement Unit (IMU) mounted under a hardhat, an ATMega Arduino attached behind the hardhat, a 120Hz binocular Pupil Labs eye tracking glasses (ETG)~\cite{Kassner2014Pupil:Interaction} and a ZED stereo camera (see Figure~\ref{fig:HW_setup}). To ensure its applicability in a wide variety of domains, the Gaze-in-Wild dataset provides easy access to depth of the real world stimulus calibrated from the person's FoV. Contrary to a two IMU system,~\cite{Tomasi2016MobileStudies} we chose a single IMU system to avoid using a body worn device since many applications of eye tracking are predominately head-mounted. The hardware setup weighed 700 gms (excluding laptop weight), which is similar to previous setups.~\cite{Barnes1979Vestibulo-ocularTargets.} To reduce slippage, the hardhat was equipped with an adjustable knob to tighten its hold on a subject's head.

\subsubsection{Pupil labs eye tracking glasses (ETG)}
\label{ETG_info}
Binocular eye trackers usually contain two eye cameras and a single world camera (which captures the scene in front of a person). Eye tracking solutions require some form of eye feature (derived from images captured from the eye camera) to Point of Regard (PoR - pixel position on the world camera) mapping to provide an accurate gaze estimate. This process is also known as eye tracker calibration. Mapping functions often vary from polynomial regression to multi-layer perceptron regression. Despite calibration, angular error tends to remain low near to the calibration region and increases radially outwards. Furthermore, there exist many sources of error which degrade the quality of gaze tracking,~\cite{Kassner2014Pupil:Interaction} particularly in unrestrained settings. The Pupil Labs eye tracker estimates the approximate center of eye ball rotation and a 3D pose of the pupil (modeled as a 3D disc). This enables the extraction of 3D gaze vectors with respect to the Eye-In-Head (EiH) coordinate system $C_{E}$. The Pupil Labs software (version number 1.8) also provides a confidence value for each gaze sample which can be interpreted as a reliability measure. It is calculated as a ratio of the number of support pixels by the number of pixels on the ellipse fit of an imaged pupil. Support pixels are the edge points within a threshold distance away from the pupil ellipse fit. All gaze samples with confidence below 0.3 were discarded from analysis.

\subsubsection{Inertial Measurement Unit (IMU)}
The MPU-6050 is a low cost 6-axis IMU that integrates a 3-axis accelerometer and a 3-axis gyroscope to estimate its pose relative to its initial position at the onset of data acquisition. The IMU is connected to an Arduino placed behind the hardhat, which in turn, is connected to the laptop backpack. The Digital Motion Processor inside the IMU provides its pose estimate at 100$Hz$. The I$^2$Cdevlib open source library was used to extract information from the IMU.~\cite{i2cdevlib} Pose estimates using an IMU sensor are known to drift due to error accumulation making it necessary to offset the IMU regularly to avoid drift in orientation measurements. Calibrating the IMU's offset at the beginning of data collection and fine tuning during post processing ensures accurate head pose within 7$^\circ$ ($\sigma=8.34^\circ$) of error for short recordings. Longer recordings may incur significant error in pose estimates unless externally corrected or reduced using a secondary sensor. Frequent head turns may also lead to an increase in head pose error so it is a good practice to reset the IMU following a few head turns.~\cite{Tomasi2016MobileStudies} While we do not hinder participants mid task, pose estimates for certain recordings (marked with $\gamma$ in Supplementary Table~\ref{tbl:Dataset_status}) were manually corrected by a rotation operation before and after each heading change during post processing. Head angular drift and deviation in orientation are measured for all participants by the difference in head pose at the beginning and end of a task. We evaluated the sensor drift to be 0.021$^\circ/s$ ($\sigma$=0.035) on average. Per participant drift can be found in Supplementary Table~\ref{tbl:Dataset_status}.

\subsubsection{ZED Stereo camera}
The ZED stereo camera provides a 1080p point cloud at 30$Hz$ which is calibrated and mapped onto the ETG coordinate system $C_{E}$ from its own coordinate system $C_{Z}$. We found the error in depth measurement to be proportional to the distance under consideration. The euclidean 3D error was found to be less than 0.5$m$ at a distance of $\sim$10$m$ (beyond that is considered to be infinity), which is in agreement with other independent analysis.~\cite{Ortiz2018DepthStereolabs}

\subsection{System calibration}
Prior to data collection, we instructed the participants to perform two calibration routines before each task.

\begin{figure}
    \centering
    \includegraphics[width=0.7\linewidth]{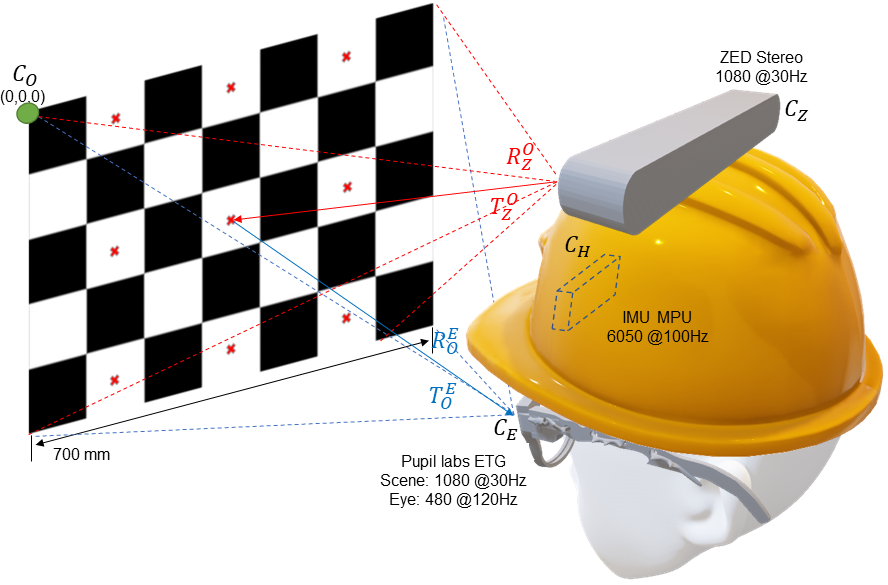}
    \caption{Checkerboard pattern placed in front of a participant during calibration phase. The red cross marks are used to calibrate the eye tracker in routine 1. The checkerboard corners are used for a 2-way multiview calibration between the ZED stereo camera $C_Z$ and the eye tracker world camera $C_{E}$. ($R^O_{Z}$, $T^O_{Z}$) $\rightarrow$ Transformation needed to move from $C_Z$ to the calibration chart's coordinate system, $C_O$. ($R^E_{O}$, $T^E_{O}$) $\rightarrow$ Transformation needed to move from $C_O$ to $C_E$}
    \label{fig:Coordinate_Transform}
\end{figure}

\textbf{Routine 1} - In the first routine, participants were asked to look at red markers placed in alternating boxes on a modified checkerboard chart (see Figure~\ref{fig:Coordinate_Transform}). The purpose of this routine was to calibrate the Pupil Labs eye tracker.

\textbf{Routine 2} - In the second routine, participants were asked to maintain a comfortable head pose while fixating on one of the calibration points. They were then asked to move their heads horizontally or vertically while maintaining fixation at that point, thus inducing a vestibular ocular response. This routine performed a system calibration by aligning all hardware components to a common world coordinate system.

\subsubsection{Pupil Labs eye tracker calibration}
\begin{figure}
    \centering
    \includegraphics[width=0.5\linewidth]{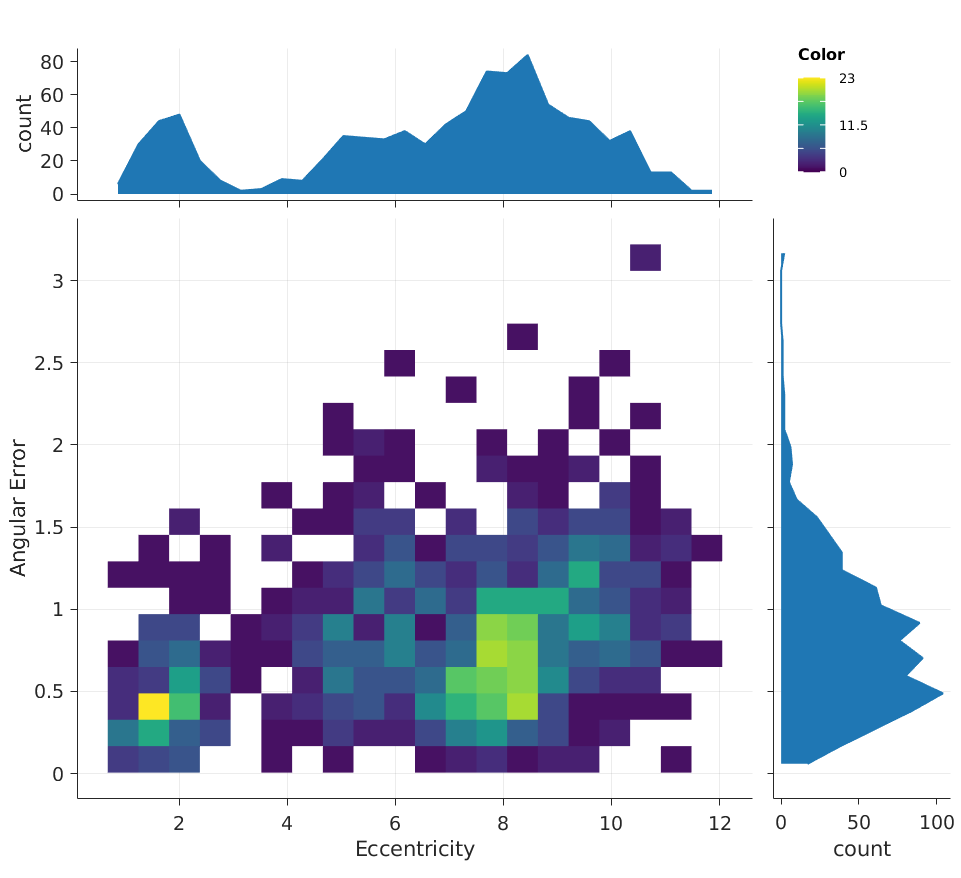}
    \caption{Eye tracker accuracy vs eccentricity from the center of the calibration pattern. Color scale indicates the number of calibration samples from all subjects.}
    \label{fig:ETG_acc}
\end{figure}

 Routine 1 is used to calibrate the eye tracker. Each marker-fixation pair is used to fit a polynomial mapping function between the 2D pupil position and 2D gaze POR. The angular gaze error reported (see Figure~\ref{fig:ETG_acc}) is measured by $\measuredangle (k_e^{-1} P_{x}, k_e^{-1}P_{c})$, where $P_{x}$ and $P_{c}$ are the homogeneous coordinates of a calibration point and gaze PoR. $k_e$ is the intrinsic matrix of the ETG world camera. We evaluated the calibration accuracy to be within 1$^\circ$ of error within 10$^\circ$ from the center of the calibration pattern. Individual participant eye tracker calibration error can be found in \ref{tbl:Dataset_status}. The ETG eye camera has manual focus lenses which were readjusted for every participant to ensure sharp visual features.

\subsubsection{Temporal alignment}
Each individual component of our system has a fixed temporal offset from each other. This temporal offset is removed using normalized cross-correlation of the angular velocity traces between the IMU, ETG and the ZED stereo camera. Since the ZED camera utilizes visual odometry to derive a pose estimate, it is not uncommon to observe a poor pose estimate during the VOR calibration routine. In those situations, we tracked the checkerboard corners in the ZED and ETG world camera to derive a velocity estimate for each corner point. In the absence of ZED pose information, these velocity estimates were used to compute the offset between ETG and ZED using cross-correlation as described in the next section. It should be noted that there exists an inherent latency between head and eye movements during a VOR \cite{Vercher1991Eye-headFixation.}. However, we remove all latency while correcting for temporal offsets (including biological latency).

\subsubsection{ETG-IMU calibration}
Initially, the IMU and ETG are defined in their own respective coordinate systems, $C_H$ and $C_E$. When participants were fixated at a point on the calibration chart during Routine 2, their eye and head pose was defined as the Z axis of our new world coordinate system $C_W$. The IMU is placed approximately 1-2 $cm$ above the cyclopean gaze origin (an imaginary point midway on the line joining both eye centers). Instead of correcting for translation offset (which can vary by subject), we choose to align $C_H$ and $C_E$ to $C_W$ solely using rotation matrices $R_H^W$ and $R_E^W$. These matrices were initially derived using vector rotations and manually fine tuned until the coordinate systems were satisfactorily aligned (Gaze-in-World (GiW) velocity, \ie the head compensated cyclopean eye velocity is minimized). Once the head and eye orientation are defined in $C_W$, we rotate the EiH vector using the updated head pose to obtain the GiW vector.

\subsubsection{ETG-ZED calibration}
Calibrating the ETG and ZED is required to register the depth point cloud from ZED's coordinate system $C_Z$ to the ETG scene camera $C_E$, to obtain calibrated depth values of the visual field. The visual field is defined from the center of the world camera, hence we choose to superimpose the depth map onto the world imagery. Since the distance between each checkerboard corner point is known, we can produce a grid of corner points in world units ($mm$) defined in the checkerboard coordinate system $C_O$. This grid can be aligned and projected on $C_E$ and $C_Z$ using extrinsic parameters ($R, T$). Corner points extracted from time synced ETG world and ZED left camera images were used to find $R$ and $T$. The extracted image points and the checkerboard grid are related using $x_Z = k_Z(R^Z_O X_O + T_O^Z)$ and $x_E = k_E (R^E_O X_O + T_O^E)$. Here, $X_O$ is the 3D checkerboard grid defined in $C_O$. $k_Z$ and $k_E$ are the left ZED and world camera intrinsic matrices. For detailed information regarding this process, we refer the reader to single camera calibration, part 1, multiview geometry by Hartley \etal~\cite{Hartley:2003:MVG:861369} The transformations required to align $C_Z$ to $C_{E}$ can be derived as $R^E_Z = R^E_{O}{R^Z_{O}}^{-1}$ and $T^E_{Z} = T^E_{O} - R^E_Z T^Z_{O}$, which are used to transform the depth point cloud from $C_Z$ to $C_E$. Once we have an aligned depth map, we trace a ray from the ETG world camera center to a subject's PoR and intersect it with the transformed point cloud to derive a 3D PoR in $mm$.

\subsection{Operations}
\label{ops}
In this paper, all absolute angular velocity measurements are calculated using a modified Two-Point Central Difference algorithm (2-P)~\cite{Bahill1982FrequencyAlgorithm}. The angular velocity $\omega_v$ can be derived as $\nicefrac{\delta \theta}{\delta t}$, where $\delta \theta$ is given by $\measuredangle (v_{n+1}, v_{n-1})$. Here, $v_n$ is a normalized unit direction vector while $t_n$ is the timing associated with sample $n$. $\delta \theta$ is the angular displacement within the elapsed time. For a fixed sampling rate $f_s$, $\omega_v = f_s \nicefrac{\delta \theta}{2}$.

Pupil tracking it usually performed in the near infrared because the human iris, regardless of color in the visible spectrum, reflects well in the near infrared. This ensures adequate contrast between the iris and the pupil, which is dark when illuminated off axis. However, noise may be introduced while tracking the pupil due to many external and internal factors such as varying illumination conditions, algorithmic artifacts, lack of contrasting eye features, occluded pupils etc. These artifacts may result in high frequency noise in the pupil positional signal. Since we are imaging the eye with a sampling frequency of $f_s$, frequencies higher than the Nyquist frequency ($f_{n}$ = $f_s/2$) are aliased into our signal as noise. To avoid aliasing, we introduce a low pass filter to suppress all frequencies higher than $f_n$ (Kaiser window, cut-off: 58$\pm$2 $Hz$). Furthermore, the 2-P central difference algorithm results in gain suppression near $f_n$ without exhibiting phase shifts as opposed to other non-symmetric techniques wherein signal delay is not constant. Phase offsets due to anti-aliasing filters are removed by performing Zero-Phase filtering.~\cite{Chaparro2015SignalsEdition} To further reduce noise, we utilize Bilateral filtering~\cite{Paris2007AApplications} since it provides an optimal trade-off between noise removal while maintaining characteristics of eye movements (such as preserving peak saccade velocity). Non-adaptive techniques such as Gaussian filtering suppress saccade velocity peaks while increasing their duration and potentially produce misleading characteristics which could lead to misinterpretation of eye movements. The optimal parameters for bilateral filtering were empirically derived (window length ~50$ms$, $\sigma_t$ = 18$ms$, $\sigma_r$ = 8.75$^\circ/s$). The azimuthal and elevation velocity components are calculated using small angle approximations because of numerical stability during quadrant changes. That is, $\omega^{Az}$ = $\nicefrac{\delta \theta^{Az}}{\delta t}$. $\delta \theta^{Az}$ is approximated as $\sin{\delta \theta^{Az}}$. Small angle approximation results in 1$\%$ error in measurement at 14$^\circ$. Assuming a maximum human angular velocity of 900$^\circ/s$, the upper limit for human angular displacement cannot exceed 6$^\circ$ for 2 samples at our sampling rate of 300$Hz$, which is well within 1$\%$ measurement error.

\section{Labelling}
\label{Labelling}
\begin{figure}
\centering
  \includegraphics[width=0.8\linewidth]{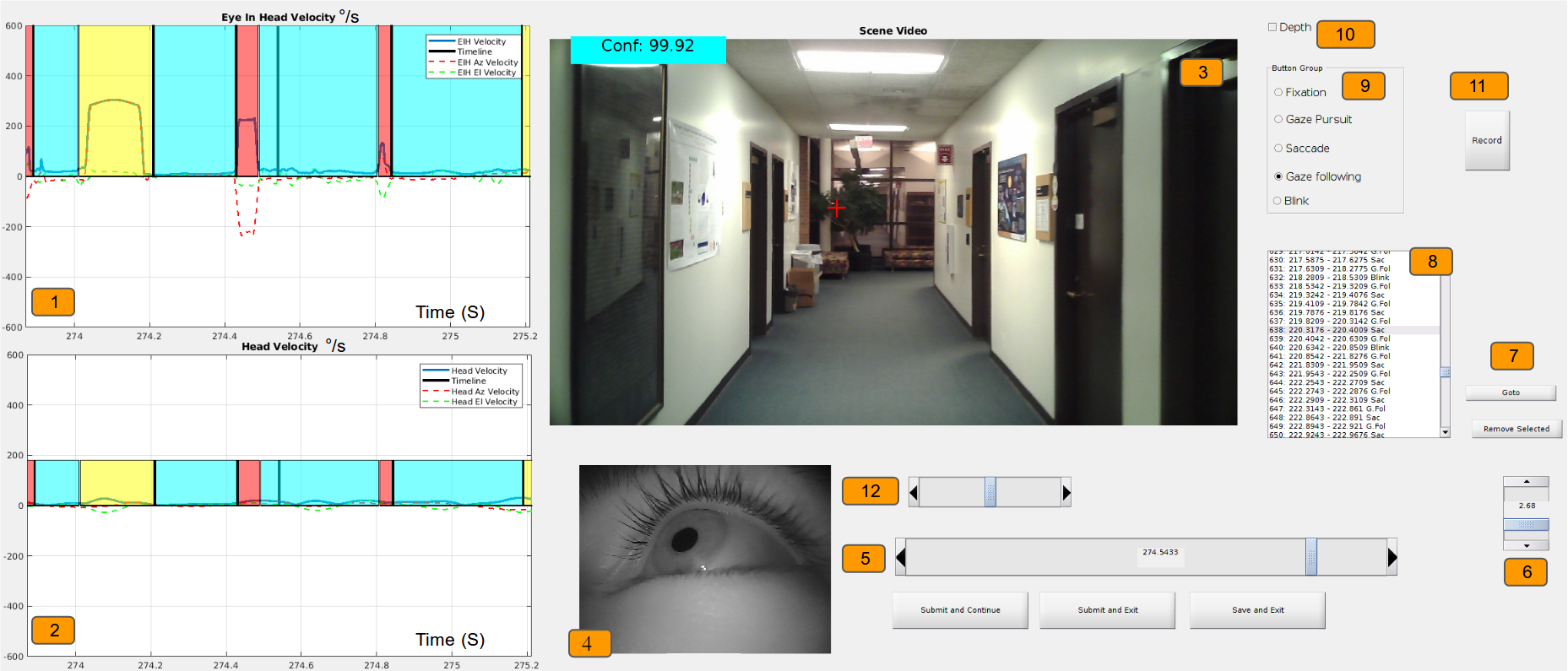}
  \caption{Custom made GUI for labelling. 1: Absolute EiH velocity with $Az$ (Azimuthal) and $El$ (Elevation) velocity traces ($^\circ/s$). 2: Absolute Head velocity with $Az$ and $El$ velocity traces ($^\circ/s$). 3: World-view overlaid with the Point-of-Regard (PoR) and confidence score. 4: Eye-view. 5: Slider to move a window through a recording temporally. 6: Slider to change the window width. 7: \textit{Go-to} and \textit{Remove button} for labelled regions. 8: Interactive list of labels in a session. 9: Radio buttons to select event type and mark across a region. 10: Toggle scene and depth view. 11: Record a 10 second clip of GUI starting at the current sample. 12: Slider to shift labels forward or backward.}
  \label{fig:Labeller}
\end{figure}

Training and evaluating a gaze event classification model requires labelling our dataset which is one of the major contributions of this work. The GW dataset was hand-labelled by five annotators who were trained to identify head-free gaze events. They produced over 140 minutes of hand-labelled head-free gaze behavior data. The dataset contains approximately 20,000 detected fixation events, 18,000 saccades, 1,200 pursuit events, and 4,000 blinks. Using a custom labelling tool (see Figure~\ref{fig:Labeller}), labellers had access to eye images, scene images with PoR cross-hair, and the individual head and eye velocity traces. Using our tool, one minute of recorded data requires 45-60 minutes of annotator time. While it is possible to develop tools that allow faster labelling\cite{Agtzidis2017InViewing}, they may bias the labeller with automated suggestive labels. Each labeller made decisions independently and they were encouraged to leave sequences where they were uncertain of the classification untouched. These sequences, along with low confidence samples (confidence below 0.3, see section \ref{ETG_info}), were treated as unlabelled and were not used to compute statistics or to train/evaluate models. While we do observe saccades as low as 15$^\circ/s$, we do not label microsaccades or post saccadic oscillations due to system accuracy limitations (head compensated gaze tremor was found to be $\mu$=$0.55^\circ/s$, $\sigma$=$0.3^\circ/s$). To provide maximum flexibility to researchers, we labelled stable fixations (caused due to tremors, drift and micro-saccades) and rVOR as a single gaze event type, stationary fixation, while labelled fixations due to tVOR and optokinetic stimulation as another gaze motion category, fixation under translation (labellers used \textit{gaze following} as a pseudonym). This enables researchers to isolate the influence of compensatory mechanisms using a variety of statistical methods.

In trials which were labelled by multiple labellers, we evaluated per-class and overall Cohen's Kappa $\kappa$ and found that human labellers exhibit an average $\Tilde{\kappa}$ of 0.72 ($\sigma$=0.0336, median=0.731). Agreement levels vary significantly by gaze event type (see Table \ref{tbl:labeller_sample_perf}, \ref{tbl:labeller_event_perf}, \ref{tbl:labeller_confmat}). Previous studies have shown that human coders exhibit a performance above 0.85 $\Tilde{\kappa}$ while classifying head fixed eye movements, with a very low inter-rater variance.~\cite{hooge2017human} While we have not managed to replicate such a high level of agreement, we can offer insights as to why. First, head-free gaze behavior is significantly more complex with a wide range of behaviors to be classified into the previously mentioned labelling scheme in Section~\ref{label_scheme}. For instance, consider classification of head-free gaze behavior while attempting to catch a ball into periods of gaze fixations, saccades and pursuit. Subjects engaged in head-free gaze pursuit for a very small portion of the ball trajectory, primarily relying on a series of fixations and predictive saccades to track the moving ball. This distinction is not straightforward and can easily be overlooked during labelling. Secondly, relying on a single source of information such as visual imagery or gaze signals could lead to incorrect coding (see Supplementary Figure~\ref{fig:example_signals}). Signal filtering and interpolation produces artifacts which may be interpreted differently by each rater.~\cite{Andersson2017OneAlgorithms} Despite the fact that we have provided multiple sources of information, it is not uncommon for a human labeller to make erroneous decisions. Lastly, while it is accepted that human coders may change their labelling strategy over time~\cite{hooge2017human} and the start and end times of coded events may vary, lack of holistic task awareness could result in data misinterpretation.

\begin{table}
\centering
\renewcommand{\arraystretch}{1.25}
{%
\begin{tabular}{|c|c|c|c|c|c|c|}
\hline
 & \multicolumn{2}{l|}{Fixational samples} & \multicolumn{2}{l|}{Gaze-pursuit samples} & \multicolumn{2}{l|}{Saccade samples} \\ \hline
 & $\mu$ & $\sigma$ & $\mu$ & $\sigma$ & $\mu$ & $\sigma$ \\ \hline
$\kappa$ & 0.72 & 0.03 & 0.68 & 0.03 & 0.75 & 0.03 \\ \hline
$p/r$ & 0.94 & 0.02 & 0.73 & 0.13 & 0.79 & 0.10 \\ \hline
F$_1$ & 0.94 & 0.02 & 0.71 & 0.04 & 0.78 & 0.03 \\ \hline
\end{tabular}
}
\caption{Sample based Cohens $\kappa$, precision/recall $p/r$ and F$_1$ score between labellers}
\label{tbl:labeller_sample_perf}
\end{table}

\begin{table}
\centering
\renewcommand{\arraystretch}{1.25}
{%
\begin{tabular}{|c|c|c|c|c|c|c|}
\hline
 & \multicolumn{2}{l|}{Fixational events} & \multicolumn{2}{l|}{Gaze-pursuit events} & \multicolumn{2}{l|}{Saccade events} \\ \hline
 & $\mu$ & $\sigma$ & $\mu$ & $\sigma$ & $\mu$ & $\sigma$ \\ \hline
$l_2$ & 12.93 & 2.2 & 13.44 & 3.13 & 13.02 & 1.94 \\ \hline
$O_r$ & 0.91 & 0.01 & 0.92 & 0.02 & 0.74 & 0.03 \\ \hline
F$_1$ & 0.86 & 0.04 & 0.71 & 0.05 & 0.89 & 0.04 \\ \hline
$\kappa$ & 0.70 & 0.09 & 0.53 & 0.08 & 0.79 & 0.08 \\ \hline
$\kappa$* & 0.54 & 0.15 & 0.44 & 0.09 & 0.61 & 0.16 \\ \hline
\end{tabular}
}
\caption{Inter-labeller event based metrics. All metrics are reported by their mean $\mu$ and inter-subject standard deviation $\sigma$. $l_2$ distance of the start and end time (expressed in $ms$) of matched events using ELC.   $O_r$ is the overlap ratio between matched events using ELC. F$_1$ score as proposed by Hooge et al~\cite{hooge2017human}. Event $\kappa$ proposed by Zemblys et al~\cite{Zemblys2018GazeNetNetworks}. Event $\kappa$* found using ELC event matching. For more information on each metric, please refer to Section~\ref{event_metric}}
\label{tbl:labeller_event_perf}
\end{table}

\begin{table}
\centering
\renewcommand{\arraystretch}{1.25}
\begin{tabular}{|c|c|c|c|c|c|c|}
\hline
                     & \multicolumn{2}{c|}{Fixational samples} & \multicolumn{2}{c|}{Gaze pursuit samples} & \multicolumn{2}{c|}{Saccade samples}  \\ 
\hline
                     & $\mu$          & $\sigma$               & $\mu$  & $\sigma$                         & $\mu$  & $\sigma$                     \\ 
\hline
Fixational samples   & 0.94           & 0.02                   & 0.02   & 0.02                             & 0.03   & 0.02                         \\ 
\hline
Gaze pursuit samples & \textbf{0.24}  & 0.11                   & 0.72   & 0.13                             & 0.03   & 0.02                         \\ 
\hline
Saccade samples      & \textbf{0.18}  & 0.08                   & 0.02   & 0.02                             & 0.79   & 0.09                         \\
\hline
\end{tabular}
\caption{Normalized sample based confusion matrix (created by normalizing the confusion matrix with the number of samples for each event type in the ground truth) across every recording with multiple labellers. Human labellers disagree the most on gaze pursuit movements.}
\label{tbl:labeller_confmat}
\end{table}

\subsection{Training labellers}
Our labelling team was trained using lectures on eye movements, gaze interpretation and eye-head coordination from the literature to thoroughly understand the labelling nomenclature used in GW. They were then asked to label a common, very small subset of the dataset that was then analyzed and discussed as a group with the authors. The labellers began manually annotating the GW dataset following this group exercise. Individual weekly meetings with the authors were set to discuss periods of uncertain data.

\subsection{Data cleaning and post-processing}
To remove erroneous labels, we adapt the approach proposed by Zemblys \etal~\cite{Zemblys2018UsingData} For our dataset, fixational events with $<$0.5$^\circ$ separation between them and within 75$ms$ of each other were combined into a single event. Fixations less than 50$ms$ and saccades greater than 150$ms$ in duration were automatically removed. Finally, labelled events with duration less than 10$ms$ were automatically removed.

\section{Error metrics}
\label{event_metric}

Evaluating the performance of automated classification systems or human labellers is not straightforward. Traditional error metrics give sample-level measurements (\eg percentage of individual samples correctly classified) and evaluate performance on a global basis, thus oblivious to the inherent structure of the data. For instance, metrics such as accuracy, precision, recall and F$_1$ score are widely used to evaluate the performance of head fixed gaze classification algorithms.~\cite{Zemblys2018UsingData, Andersson2017OneAlgorithms, Hoppe2016End-to-EndNetworks} For evaluating agreement level among labellers or classifier performance with unbalanced data (large variation in the number of samples per class), accuracy based error metrics suffer from the \textit{Accuracy Paradox}~\cite{Zemblys2018GazeNetNetworks} which means that a predictive model with high sample level scores might have a lower event prediction ability. Powers observed that symmetric kappas (\eg Cohen's kappa), which are designed for inter-rater metrics, may not be directly suitable for automated classifiers.~\cite{Powers2012TheKappa} Sample based measures fail to account for any temporal structure and may not reflect the severity of misclassifying a few, albeit structurally important, samples. Furthermore, it is more intuitive to reason in terms of correctly/incorrectly classified collections of continuous samples of the same class, or \textit{events}.

Event based metrics were designed to compensate for the limitations of sample based evaluation methods. Hoppe \etal provided the percentage of correctly classified events by comparing the samples within the bounds of each groundtruth event. The category with the highest number of samples was matched with the reference event.~\cite{Hoppe2016End-to-EndNetworks} Hooge \etal proposed a set of evaluation metrics such as the event-level F$_1$ score, the relative timing offset (RTO) and the relative timing deviation (RTD) between matched events.~\cite{hooge2017human} To compute the F$_1$ score for a particular gaze movement category, they treat every other category as a common opposite category. However, this operation removes all inter-category confusion. The first overlapping testing event of the same category as the groundtruth is considered as matched. Temporal offsets between event start and end times are calculated for all matched events, providing the added benefit of a measure for temporal alignment quality. Zemblys \etal proposed the event error rate (EER), which is a length normalized Levenshtein distance between event sequences.~\cite{Zemblys2018GazeNetNetworks} Zemblys \etal also proposed the event-level Cohen's kappa measure, an extension of the event-level F$_1$ score.~\cite{Zemblys2018GazeNetNetworks} These proposed event level metrics use the standard available measures (F$_1$, Cohen's $\kappa$) but vary in their event \textit{matching} scheme. Differing from Hooge \textit{et~al.}, Zemblys \etal proposed that a testing event with the highest overlap ratio with a groundtruth event is to be treated as a match. Note that events of differing categories may also be considered as \textit{matched}. This results in an event level confusion matrix which is used to generate an overall and per category Cohen's kappa score. Existing event level metrics improve the way we evaluate the performance of temporal classifiers but have their own individual shortcomings for varying scenarios. For instance, the majority vote method gives no penalty to unexpected short events that split longer events, and significantly influence the statistical distribution.~\cite{Hoppe2016End-to-EndNetworks,Zemblys2018GazeNetNetworks} The event level F$_1$ score also does not support multi-class evaluation,~\cite{Zemblys2018GazeNetNetworks} and the EER measure does not match events and treats all event sequences as strings. It does not consider or provide insight into temporal offsets. Furthermore, it also suffers from the \textit{Accuracy Paradox} and only returns a single value as an overall rating. Last but not least, different event-matching procedures significantly affect the RTO and RTD measurements. Zemblys \etal identified that the RTO and RTD measures will be compromised when using the largest overlapping event-matching strategy.~\cite{Zemblys2018GazeNetNetworks} Similar situations may occur when utilizing the earliest overlapping matching strategy. For example, when onset of the earliest overlapping testing event is close to the offset of a reference event. Various event based metrics are summarized in Table~\ref{tbl:event-metric-comparison}. To address some of the shortcomings of previous approaches, we devised the Event Level Cross-Category Metric (ELC) as described below.
 
\begin{table}
\centering
\begin{tabular}{|c|c|c|c|c|c|c|}
\hline
 & Matching technique & \begin{tabular}[c]{@{}c@{}}Timing \\ offsets\end{tabular} & \begin{tabular}[c]{@{}c@{}}Confusion \\ matrix\end{tabular} & Symmetric & \begin{tabular}[c]{@{}c@{}}Threshold \\ dependency\end{tabular} & \begin{tabular}[c]{@{}c@{}}Reliability of\\ timing offsets\end{tabular} \\ \hline
Majority vote~\cite{Hoppe2016End-to-EndNetworks} & Sample-level majority vote & $\times$ & $\checkmark$ & $\times$ & $\times$ & N/A \\ 
Event F1~\cite{hooge2017human} & Earliest overlapping event & $\checkmark$ & $\times$ & $\times$ & $\times$ & low \\ 
Event kappa~\cite{Zemblys2018GazeNetNetworks} & Largest overlapping event & $\checkmark$ & $\checkmark$ & $\checkmark$ & $\times$ & low \\ 
Event error rate~\cite{Zemblys2018GazeNetNetworks} & N/A & $\times$ & $\times$ & $\checkmark$ & $\times$ & N/A \\ 
ELC & Window-based matching & $\checkmark$ & $\checkmark$ & $\times$ & $\checkmark$ & high \\ \hline
\end{tabular}
\caption{Comparison of event level error metrics}
\label{tbl:event-metric-comparison}
\end{table}

\begin{figure}
\centering
\includegraphics[width=0.6\linewidth]{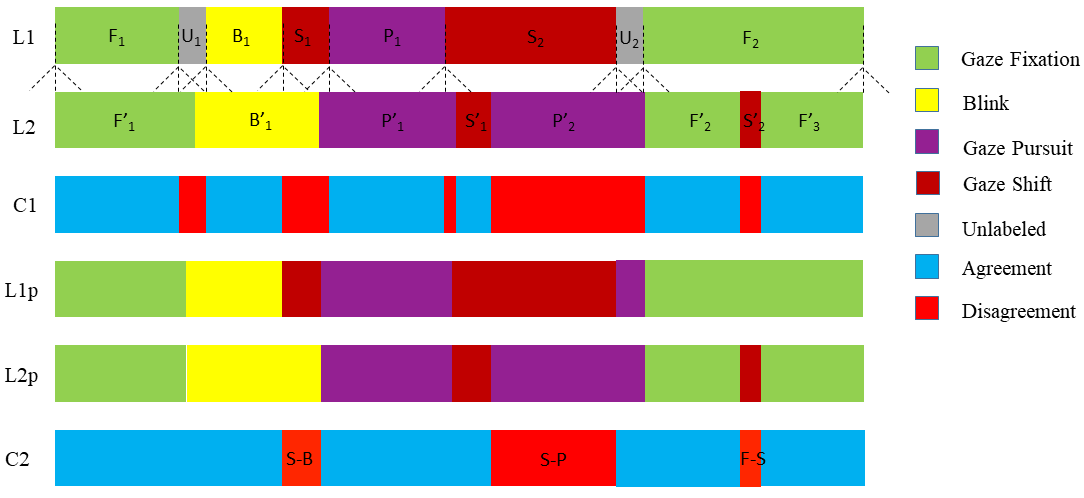}
\caption{Illustration of the ELC metric on handcrafted test and reference sequences. (L1) Labels provided by labeller 1. (L2) Labels provided by labeller 2. Colors indicate the event type and whether labellers are in agreement. Dotted lines from L1 into L2 indicate the time window used in ELC for each transition point. (C1) Direct sample-sample comparison between labeller 1 and labeller 2. (L1p, L2p) Results of applying ELC to labels provided by labeller 1 and labeller 2 respectively. (C2) Event-level comparison between labeller 1 and labeller 2. Unmatched regions are given specific labels describing the misclassification type. For example, `S-B' means that labeller 1 labelled the data as gaze shift whereas labeller 2 labelled the same data as blink.}
\label{fig:event_metric}
\end{figure}

Consider the following taxonomy:
\begin{itemize}
    \item Reference sequence - groundtruth sequence of labels.
    \item Testing sequence - predicted sequence of labels, usually the output of an automated classification process.
    \item Matched event - two events are considered matched when their start and end position roughly align in a predetermined window and meet the matching criterion (discussed below). As an example, consider sequences L1 and L2 in Figure~\ref{fig:event_metric}. All fixation events in L1 (marked in green) are considered as matched.
    \item Unmatched event - All events which do not satisfy our matching criterion are considered as unmatched. Both saccades in Figure~\ref{fig:event_metric}, are considered as unmatched.
    \item Detached event - We often find unmatched events in our ground truth which completely overlap with another test event and belong to the same gaze category. These type of events are considered to be detached. For example in Figure~\ref{fig:event_metric}, the blink in L1 (marked in yellow, the start point is matched whereas the end point has no match) is considered as a detached event. Researchers may safely consider detached events as matches per their strictness requirements and application (this operation would inflate the performance score of a classifier).
    \item Transition point - It is assumed that all event boundaries touch each other at their transition points. Transition points have samples of different gaze behavior adjacent to it. In case event boundaries do not touch, we assume the period between them to be the \textit{none} class. All entries pertaining to \textit{none}, i.e blinks and unlabelled periods, are removed from consideration. Note that all events have two transition points.
\end{itemize}

\begin{enumerate}
\item \textbf{Window-based matching:}
First, we identify every transition point in the reference and testing sequence. For every transition point in reference sequence, we extend a window of a certain size (\eg 50 $ms$) onto the test sequence and find all transition points within. The reference transition point is matched with the first (in time) testing transition point within the match window which satisfies a particular matching criterion. For onset transition points, the event type on the right should match the reference event type. Similarly, offset transition points are matched if the event type on its left matches the reference event type. An event is matched if both its start and end transition points are matched.  

\item \textbf{$l_2$ distance calculation for matched events:} 
Following window-based matching, overall timing offsets are calculated for matched events. Unlike RTO and RTD, which are calculated separately for start and end points of events, we calculate the $l_2$ distance ($\sqrt{(start_1-start_2)^2+(end_1-end_2)^2}$ where $start_1$ and $start_2$ is the start positions of two events, $end_1$ and $end_2$ are end positions of two events) for each event. The mean and standard deviation of all calculated $l_2$ distances (per class and overall) are used as indicators of alignment quality between two labelled sequences.

\item \textbf{Overlap ratio calculation for matched events:}
Since events of different categories have various ranges of duration, the severity of temporal misalignment could be different for individual event types having the same timing offset values. Therefore, we calculate the \textit{overlap ratio}~\cite{kisler2013dialect} $O_r$ ($O_r = \nicefrac{n_1 \cap n_2}{n_1 \cup n_2})$, where $n_1$ and $n_2$ are samples belonging to two matched events. The overlap ratio reflects the temporal alignment quality of two events. As with the $l_2$ distance, the mean and standard deviation are calculated and reported.

\item \textbf{Timing offsets correction:}
Once the $l_2$ distance and overlap ratio is calculated, we remove the effects of misalignment by correcting timing offsets in both sequences. This correction is applied on all matched transition points regardless of an event's match status. For each matched transition point, timing (sample index) of two points are averaged to create a single representative transition point. If the original point is shifted away from the event center, the displaced samples are assigned the event's category. Likewise, if the transition point moves inwards, the displaced samples are assigned the external event's gaze category. If the displaced samples are unlabelled in a particular sequence, they are assigned the same gaze movement class as the corresponding sequence.
\item \textbf{Event level confusion matrix:}
Comparing two labelled sequences leads to a collection of matched and unmatched events, \ie a confusion matrix, which describes inter-category event classification performance. Owing to the timing offsets correction step, event mismatches within the preset threshold are eliminated. Standard metrics such as Cohen's $\kappa$ and F$_1$ score can be derived from the confusion matrix for deeper insights or to summarize performance. 

\item \textbf{Applying previous steps in both comparing directions:}
ELC is an asymmetrical event matching technique. It can be applied twice by interchanging the testing and reference sequences to find an average performance measure along with a sense of metric agreement. For instance, if the number of detached events is higher in a particular order, it provides insight into larger proportions of event merges in the testing sequence. Inter-labeller performance is computed by applying ELC both ways but not for human-classifier evaluation.


\end{enumerate}

In Figure~\ref{fig:event_metric}, sequences L1p and L2p show the results of applying ELC to the labels in L1 and L2 respectively. The application of these rules eliminates many minor (mainly temporal) disagreements between sequences and considers only the regions of major disagreement as seen in sequence C2. Event Kappa utilizes the largest overlapping strategy to match events, which results in lower RTO and RTD scores.~\cite{Zemblys2018GazeNetNetworks} For instance, event $F_2$ in L1 gets split into two shorter events $F'_2$ and $F'_3$ by an unexpected event $S'_2$ in L2, the metric tends to match the fixation in L1 with the largest overlapping event ($F'_3$ in this case). This leads to a poor RTO and RTD measures. However, ELC considers the start and end points of $F_2$ in L1 and matches them with the start of $F'_2$ and the end of $F'_3$ respectively. $F_2$ is considered as a matched event and the testing sequence is rewarded by increasing the F/F counter in the confusion matrix. Likewise, the testing sequence is scored negatively for the offending event, $S'_2$, by increasing the F/S counter in the confusion matrix. The $l_2$ distance (functionally equivalent to RTO and RTD measurements) accurately computes the alignment quality. Interchanging L2 as the reference and L1 as the testing sequence, events $F'_2$, $S'_2$ and $F'_3$ would be considered as unmatched events and $l_2$ distances would not be calculated.

Overall, ELC provides a faithful indication of timing offsets using the window-based matching strategy. ELC is dependent on a parameter, \ie the window size. The window size indicates the system tolerance for timing offsets between ground truth and testing events. Since it's easier to identify the start and end points of gaze shifts as compared to other types of gaze events, different window sizes for gaze shift related events ($\pm25 ms$) and non gaze shift related events ($\pm35 ms$) are used. Researchers may consider using larger window sizes for situations wherein event onset and offsets conditions are relaxed.



\section{Machine learning for gaze event classification}
\label{classification}

We trained two standard machine learning models for gaze event classification: a moving window based method and a recurrent neural network (RNN). The input to both classifier models is a sequence of temporally discrete sensor data vectors, i.e., $\mathcal{D}= \left\{ \mathbf{x}_1, \mathbf{x}_2 , \ldots, \mathbf{x}_n , \ldots , \mathbf{x}_T \right\}$, where $\mathbf{x}_n \in \mathbb{R}^d$ and $n$ is the current time step. As described in Section~\ref{classification:features}, these data vectors contain information from the IMU and eye tracker. For both models, we merge fixations when stationary and fixations under translation into a single gaze fixation class (see Section~\ref{label_scheme} and Section~\ref{Labelling}).


\subsection{Classification models}
\label{classification:models}
The moving window model classifies a gaze sample at time $n$ by aggregating information from a window of data vectors adjacent to $\mathbf{x}_n$, i.e.,  $\mathbf{w}_n = W\left(\mathbf{x}_{n-s}, \ldots, \mathbf{x}_{n}, \ldots \mathbf{x}_{n+s} \right)$, where the vector of window features $\mathbf{w}_n \in \mathbb{R}^g$ is computed using a window size of $2s + 1$ samples and the function $W \left( \cdot \right)$ computes the windowed feature vector. We chose the random forest (RF) classification algorithm since it works well for low-dimensional data, and our framework resembles state-of-the-art gaze event algorithms for controlled 2D environments.\cite{Zemblys2018GazeNetNetworks} RF is an ensemble learning method wherein multiple decision trees are trained on a subset of samples and their feature space.\cite{Breiman1999RandomForests} A RF is easy to train and they are robust to noise and over-fitting, which are common problems for decision trees. For gaze classification in 2D controlled environments, Zembyls \etal showed that RF performed well with only 16 trees and 10-dimensional features up to a 200 ms window.\cite{Zemblys2018GazeNetNetworks} In our experiments, we use 40 trees, a minimum leaf size of 15, and we use $\sqrt{g}$ randomly selected features per tree where $g$ is the number of features for a given window size.  For training the window-based RF classifier, we remove duplicated $\mathbf{w}$ vectors but scale its average confidence measure by the number of duplicates found. This step is effective in reducing memory usage and training time by reducing the size of the training set. For the test set, no duplicates are removed.

Rather than using explicit windows, the RNN model operates on the velocity data stream, i.e the absolute, azimuthal and elevation velocity (see Section~\ref{ops}). We use two variations of the RNN model. Our one directional forward RNN model (fRNN) classifies the gaze at time $n$ using only past and present information, i.e., $F \left( \mathbf{x}_1 , \mathbf{x}_2, \ldots,  \mathbf{x}_n \right)$. This model would be especially useful for real-time gaze prediction. For offline processing, we also use a bi-directional RNN (biRNN) that has past, present, and future information as input, i.e., $F \left( \mathbf{x}_1 , \mathbf{x}_2, \ldots,  \mathbf{x}_n, \ldots \mathbf{x}_T \right)$.  Both models are implemented with gated recurrent units (GRUs),\cite{Chung2014EmpiricalModeling} which can handle longer-term dependencies than simple RNNs. A similar approach was used by the GazeNet architecture,\cite{Zemblys2018GazeNetNetworks} which used an RNN to classify events in a controlled 2D environment.

\begin{figure}
\centering
\includegraphics[width=0.5\linewidth]{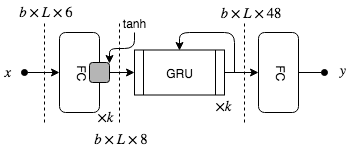}
\caption{Bidirectional recurrent network model architecture. The model takes the absolute, azimuthal and elevation eye and head velocity (6 features) as its input, passes through $k$ fully connected feature extraction layers. These features are fed into a stack of $k$ GRU layers which learn temporal patterns to classify a sample $x_t$. The forward variant (fRNN) outputs a 24 dimensional vector instead of 48 before being reduced to 3 at the final FC layer.}
\label{fig:recurrent_arch}
\end{figure}

The input to our model is a subset of windowed features $W$. Specifically, the model accepts the absolute, azimuthal and elevation EiH and head velocity as input. Multiple sequences, \textit{b}, are stacked into a single batch of data. All sequences were padded with zeros to be of the same length as the longest sequence present in the batch, \textit{L}. This $b \times L \times 6$ dimensional data passes through $k$ fully connected layers which generates a nonlinear representation of EiH and head velocity. Extracted features are fed into a stack of $k$ GRU (see Figure~\ref{fig:recurrent_arch}) units which learn to associate temporal patterns with a type of gaze behavior. We use cross-entropy loss and train the network using ADAM\cite{Kingma2014Adam:Optimization} for 1000 epochs with a learning rate of 0.01, which we reduced linearly as the training performance improved. We experimented with the number of recurrent and linear layers and found $k$=3 worked best.


\subsection{Input features}
\label{classification:features}
The $\mathbf{x}_n$ features consist of normalized EiH vectors $v_e$ and head vectors $v_h$ concatenated together. For the window-based RF classifier, for each time step $n$, we extract the following set of handcrafted features from a window of size $2s+1$ around the $n$-th sample:
    \begin{enumerate}
        \item \textbf{Mean EiH and head angular distance}: $\Delta \theta_{e}$, $\Delta \theta_{h}$. Angular distance in degrees between the mean EiH/head vector of $s$ samples before and after the current sample of interest, $x_n$.
        \item \textbf{Deviation in EiH and head velocity}: $\sigma_e$, $\sigma_h$. Standard deviation of absolute EiH and head angular velocity.
        \item \textbf{Confidence}: We supply the confidence measure (see Section \ref{ETG_info}) to our classifiers as weights for each sample. High confidence and duplicate samples are assigned larger weights.
    \end{enumerate} 
    
We also aggregate velocity measurements from every sample in the window. The velocity measurements are the absolute EiH $|\omega_e|$ and head velocity $|\omega_h|$ (angular velocity extraction has been described in Section~\ref{ops}), azimuthal EiH and head velocity $\omega^{Az}_e$, $\omega^{Az}_h$ and elevation EiH and head velocity $\omega^{El}_e$, $\omega^{El}_h$. All velocity measurements are expressed in $^\circ/s$. Azimuthal and elevation velocity contain directional information such that a positive sign indicates a clockwise rotation and vice versa. Assimilating features as a time series results in a $g$ dimensional window feature vector, where $g$ = $4(2s + 1) + 6$. The full window feature vector is given by $\mathbf{w}_n = \big(\big[|\omega_e|,\ |\omega_h|,\ \omega^{Az}_e,\ \omega^{Az}_h,\ \omega^{El}_e,\ \omega^{El}_h \big]_{n-s}^{n+s},\ \Delta \theta_{e},\ \Delta \theta_{h},\ \sigma_e,\ \sigma_h \big)^T_n$, where $[ \ * \ ]$ stands for aggregation.

\subsection{Results}
\begin{figure}
\centering
\includegraphics[width=\linewidth]{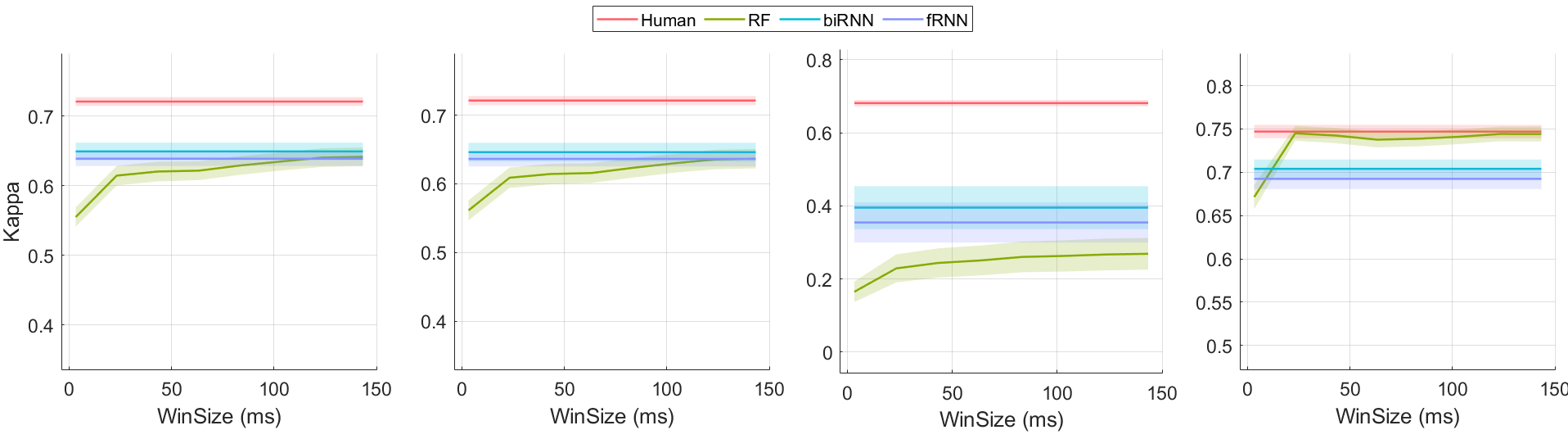}
\caption{Sample level performance metrics. All performance curves are centered around their mean, $\mu \pm$ standard error. Left - Overall $\kappa$ score. Inner left - Gaze fixation $\kappa$ score. Inner right - Gaze pursuit $\kappa$ score. Right - Saccade $\kappa$ score. Please note the varying y-limits to accentuate the difference in performance. RNN uses memory to encode temporal patterns, and hence the RNN architectures are represented as horizontal lines as they do not operate in window sizes. We would like to highlight that all window sizes are in the velocity domain. Window sizes in angular domain can be derived by adding 10 $ms$ (please refer to Section \ref{ops}).}
\label{fig:sample_kappa}
\end{figure}

The two classifiers are assessed using \textit{leave-one-out} cross validation by testing on a single person's data but training the model on remaining subjects, and comparing their average performance to human labellers. Our analysis is done using both sample and event level metrics (see Section~\ref{event_metric}). Classifier output is not evaluated during blinks or for unlabelled data points. As the window size increases, RF gains increasing temporal awareness which results in higher $\kappa$ performance with diminishing returns. It can be observed in Figure~\ref{fig:sample_kappa} that RF arrives at an asymptotically improving performance with a window size of 30 $ms$ and above. Individual $\kappa$ scores for each gaze class reveals that all classifiers find it difficult to distinguish gaze pursuits. Overall, sample based metrics convey that RF with a large window size outperforms RNN for detecting saccades but performs poorly on gaze pursuit samples.

\begin{table}
\centering
\renewcommand{\arraystretch}{1.25}
\begin{tabular}{|c|c|c|c|c|}
\hline
 & $\kappa$ & G.Fix $\kappa$ & G.Pur $\kappa$ & Sac $\kappa$ \\ \hline
RF & 0.64 & 0.64 & 0.27 & \textbf{0.74} \\
fRNN & 0.64 & 0.64 & 0.35 & 0.69 \\
biRNN & 0.65 & 0.65 & \textbf{0.40} & 0.70 \\ \hline
Human & 0.72 & 0.72 & 0.68 & 0.75 \\ \hline
\end{tabular}
\caption{Sample based Cohen's Kappa score $\kappa$ for each optimized classifier.}
\label{tbl:sample_perf}
\end{table}

\begin{table}
\centering
\renewcommand{\arraystretch}{1.25}
\centering
\begin{tabular}{|c|c|c|c|c|c|c|c|c|}
\hline
 & G.Fix F$_1$ & G.Pur F$_1$ & Sac F$_1$ & EER & G.Fix $\kappa$ & G.Pur $\kappa$ & Sac $\kappa$ & Overall $\kappa$ \\ \hline
RF & 0.80 & 0.32 & 0.85 & 0.17 & 0.61 & 0.13 & 0.71 & 0.47 \\
fRNN & 0.82 & 0.34 & 0.86 & 0.16 & 0.64 & 0.24 & 0.70 & 0.50 \\
biRNN & \textbf{0.84} & \textbf{0.42} & 0.86 & \textbf{0.14} & \textbf{0.67} & \textbf{0.30} & \textbf{0.72} & \textbf{0.54} \\ \hline
Human & 0.86 & 0.71 & 0.89 & 0.14 & 0.70 & 0.53 & 0.79 & 0.62 \\ \hline
\end{tabular}
\caption{Metrics based on various event matching techniques proposed by others. Event based F$_1$ score proposed by Hooge et al.~\cite{hooge2017human} Event Error Rate (EER) proposed by Zemblys et al.~\cite{Zemblys2018GazeNetNetworks} Event and overall $\kappa$ scores calculated using Zemblys et al.~\cite{Zemblys2018GazeNetNetworks}}
\label{tbl:event_perf}
\end{table}

\begin{table}
\centering
\renewcommand{\arraystretch}{1.25}
\begin{tabular}{|c|c|c|c|c|c|c|c|c|c|c|c|c|c|}
\hline
 &  & \multicolumn{4}{c|}{G.Fix} & \multicolumn{4}{c|}{G.Pur} & \multicolumn{4}{c|}{Sac} \\ \hline
 & Overall $\kappa$ & $\kappa$ & $O_r$ & $l_2$ & $l_2$ $\sigma$ & $\kappa$ & $O_r$ & $l_2$ & $l_2$ $\sigma$ & $\kappa$ & $O_r$ & $l_2$ & $l_2$ $\sigma$ \\ \hline
RF-144 & 0.47 & 0.43 & \textbf{0.93} & \textbf{13.1} & \textbf{2.51} & 0.07 & 0.89 & 14.43 & 4.72 & \textbf{0.59} & \textbf{0.76} & \textbf{12.89} & \textbf{2.48} \\
fRNN-3 & 0.42 & 0.39 & 0.92 & 15.17 & 3.81 & 0.15 & \textbf{0.92} & \textbf{12.69} & 5.34 & 0.5 & 0.69 & 15.02 & 3.75 \\
biRNN-3 & \textbf{0.47} & \textbf{0.45} & 0.92 & 15.11 & 3.47 & \textbf{0.24} & 0.91 & 12.95 & \textbf{4.16} & 0.53 & 0.7 & 14.86 & 3.49 \\ \hline
Human & 0.56 & 0.54 & 0.91 & 12.93 & 2.2 & 0.44 & 0.92 & 13.44 & 3.13 & 0.61 & 0.74 & 13.02 & 1.94 \\ \hline
\end{tabular}
\caption{Standard metrics derived from the ELC confusion matrix. $O_r$ is the overlap ratio between matched events. $l_2$ distance between matched event start and end times and their standard deviation $l_2-\sigma$ in $ms$. $l_2$ and $l_2-\sigma$ are similar to RTO and RTD metrics proposed by Hooge et al.~\cite{hooge2017human}}
\label{tbl:event_perf_elccm}
\end{table}

We report event based metrics and observe that biRNN outperforms RF on all measures. Interestingly, event F$_1$ and event $\kappa$ scores computed using Zembyls \etal shows an increase in saccade classification performance (see Table~\ref{tbl:event_perf}) for biRNN over RF. However, this increase is not reflected using sample based metrics (Table~\ref{tbl:sample_perf}) or event $\kappa$ computed using ELC (Table~\ref{tbl:event_perf_elccm}). Notably, RF outperforms RNN based methods in $l_2$ scores, indicating a better ability to produce tighter fits around saccades (see Table~\ref{tbl:event_perf_elccm}). Overall, gaze pursuit classification baselines fall short on human level performance but the results are consistent with the difficulty in classifying pursuit movements over other gaze movement types in general.~\cite{Pekkanen2017ARegression,Karpov2013,Santini2016} The biRNN model produces higher number of detached events (G.Fix: 0.2, G.Pur: 0.19, Sac: 0.04) as compared to RF (G.Fix: 0.07, G.Pur: 0.01, Sac: 0.11) which indicates that a larger number of ground truth events completely overlapped with events of the same category but their transitions did not fall within the matching window. Since it is debatable if these detached events can be considered as matches, we omit them from all measures to avoid inflating scores.

\subsection{Ablation study}
To understand the role of each feature, we systematically removed essential components from the best performing model (biRNN with 3 FC and GRU layers). The input to biRNN comprises of absolute EiH/head velocity, azimuthal and elevation EiH/head velocity. This generates a signal with 6 features. Please note that azimuthal and elevation velocity store relative direction information between the eye and head (-ve sign means an anticlockwise rotation). By comparing different conditions using sample based $\kappa$ score, we highlight the essential components required for head-free gaze classification in Table~\ref{tbl:ablation}. For a detailed comparison using all metrics, please refer to Supplementary Table \ref{tbl:ablation_full}.

\begin{table}
\centering
\renewcommand{\arraystretch}{1.25}
\begin{tabular}{|c|c|c|c|c|}
\hline
Cohen $\kappa$          & Overall & G.Fix & G.Pur & Sac \\ \hline
biRNN (only eyes)     & 0.62    & 0.61  & 0.29  & 0.70 \\
biRNN (only absolute) & 0.64    & 0.63  & 0.38  & 0.70 \\
biRNN                 & \textbf{0.65}    & \textbf{0.65}  & \textbf{0.40}   & 0.70 \\ \hline
\end{tabular}
\caption{Sample based $\kappa$ score after removing either head movement information or directional information.}
\label{tbl:ablation}
\end{table}

As expected, the performance of biRNN with EiH information (only eyes) did not vary while detecting gaze fixations and saccades, but drops by 25$\%$ (0.40 $\rightarrow$ 0.29) while detecting pursuit events. Interestingly, a few pursuit events were still detected despite the lack of head movements. This indicates that head-free eye movements during pursuit behavior show a varied velocity pattern than gaze fixations and can be differentiated without any knowledge of head motion. We also observe that there is a minor loss in performance when we remove azimuthal and elevation components. This highlights that absolute velocity information alone can provide reasonable certainty for classification.

\section{Discussion}
The main purpose of this work was to build the first dataset of labelled gaze movements collected during natural behavior `in the wild' (outside of the laboratory), to have multiple labellers manually label the gaze events in the dataset, and to showcase the performance of two standard temporal classification techniques, Random Forest and Recurrent Neural Networks, using some common evaluation metrics. To overcome incorrect inter-event timing offsets observed in existing metrics, we introduce the ELC metric. The usefulness of a classifier lies in its ability to generalize in unseen circumstances. Hence, all our baseline performances are evaluated using the \textit{leave-one-out} approach, wherein a classifier is tested on a single subject's data while trained on the rest. Despite the fact that there is variability among human labellers, there is as yet no other choice, so we rely here on their labels as the gold standard. To improve upon existing event metrics and provide a reliable measure of alignment quality, we devised a new event matching technique, ELC, which matches events based on their transition points. ELC provides some control on evaluation strictness by identifying events which belong to the same category but are not temporally aligned due to event fragmentation.

\subsection{Lower gaze pursuit classification performance by classifiers}
The best performing classifier for gaze pursuits is 35$\%$ lower than the average human level performance (sample $\kappa$: 0.62 $\rightarrow$ 0.40) whereas fixation and saccade performance achieves near 90$\%$ of human performance (sample $\kappa$ G.Fix: 0.65$\rightarrow$0.72, Sac: 0.70$\rightarrow$0.75). While pursuing moving targets, we observed that participants seamlessly interchanged between fixational and pursuit movements. The distinction between these movements are difficult to observe, especially during low velocity conditions because small angular errors in orientation measurements (a phenomenon common with IMUs) could result in misinterpretation without additional context for consideration (such as scene imagery with overlaid gaze PoR), a modality currently unavailable to our classifiers. Distinction between gaze fixation and pursuit events is further compounded when the head tracks a moving target or makes anticipatory movements but gaze remains stable at a fixation point. This motion elicits a signal similar to VOR, but if we rely purely on visual inspection then these events can easily be confused with pursuit motion. Situations such as these, combined with minor orientation errors, largely contribute to fixation/pursuit confusion seen in Table~\ref{tbl:sample_perf}.

\subsection{Head tracking: A pursuit or fixation?}
Previous research has shown that the head \textit{tracks} a moving target (in our case the ball) while the eyes \textit{predict} the ball location using predictive saccades.\cite{Mann2013TheBall} We find numerous instances of gaze shifts to known targets where head movements precede eye movements in an anticipatory manner~\cite{Daemi2015kinematicmodel} to ensure that upcoming eye movements do not deviate too far from the relatively tight distribution seen in Figure~\ref{fig:head&eye_dist}. Participants frequently showed tracking behavior with the head and predictive or catch-up motion with the eyes during early phase of the ball trajectory. This behavior is usually followed by gaze pursuits during the next phase, i.e. the ball height is peaked and its projected retinal velocity is low. Following the peak phase, participants made predictive saccades to their hand for successful ball interception. GW also captures instances where the head \textit{catches-up} to the fixation location while maintaining a strict coupling with the ball trajectory. While some may argue that head tracking of a moving object constitutes a pursuit motion, we instructed labellers to mark those sequences as fixations because the signals are identical to a VOR (please refer to Supplementary Figure~\ref{fig:example_signals}).

\subsection{Head and Eye tracking can have different coordinate systems}
Based on the ablation study, we observed that providing only the absolute velocity information achieved almost the same performance as biRNN-3, our best performing model. Interestingly, it highlights that for a slight drop in performance, future end-end classification frameworks may perform reasonably well if they simply provide unaligned eye and head motion information. While gaze fixations and saccades are distinctly identifiable using only eye-in-head (EiH) information, pursuit movements would be difficult to differentiate with a fixation without head movement information. As a sanity check, we also verified that the presence of a head tracking device improves classification of head-free pursuit movements by up to 25$\%$ as opposed to without head movement information (sample $\kappa$: 0.29 $\rightarrow$ 0.40). It is interesting to note that despite removing head movements, the RNN classifier is still able to identify a few pursuit events which indicates that they demonstrate different EiH velocity statistics as fixations (for more information, please refer to Supplementary Table~\ref{tbl:ablation_full}).

\subsection{Gaze-in-world information for classification}
We include head pose as an input modality for the classifiers. While it is possible to classify the gaze-in-world signal, which is the head compensated eye-in-head signal, we wanted to train algorithms which could directly capture eye and head movement dynamics along with classifying it. For instance, we often find gaze pursuit events which are dominated either by head or eye movements, a distinction which would be lost when classifying gaze-in-world information.

\subsection{General limitations}
Given limitations in current technology, it is unavoidable that tracking head position using a low cost IMU will accumulate error over time. All task duration were $\sim$3 minutes long and the error in orientation at the start and end of a recording was found to be 7$^\circ$ on average (see Section~\ref{Methods:HW}). While this error affects the absolute velocity component by a very small margin (0.04$^\circ/s$ on average), it leads to unwanted shifts in the azimuth and elevation velocity component (see Supplementary Figure~\ref{fig:example_signals}). Despite the use of a ratcheted head strap, this error accrues, in part, due to slippage of the helmet on the head, which will cause a misalignment of the helmet-mounted ZED stereo camera and the Pupil Labs eye tracking glasses (see Section~\ref{Methods}). Future work might further reduce slippage through using software correction, such as the estimation of rotational slip on a frame-to-frame basis by matching visual features in the stereo camera and Pupil Labs world camera imagery, or through the fusing visual pose estimates with IMU data, as is commonly used in simultaneous localization and mapping.

\subsection{Limitations of event-based metrics}
Although event level error metrics give researchers a better idea of the actual performance of automated classifiers or agreement level between labellers, existing event level metrics suffer from various drawbacks. The majority vote metric by Hoppe \etal remains agnostic to the testing sequences' structure. It does not penalize during event fragmentation caused by unexpected short events in the testing sequence.~\cite{Hoppe2016End-to-EndNetworks,Zemblys2018GazeNetNetworks} Moreover, this metric could be biased by the distribution of samples. Event level F$_1$ score does not work well in multi-category scenario~\cite{Zemblys2018GazeNetNetworks} and gives out unreliable RTO and RTD. EER does not provide any measure of alignment quality and suffers from the \textit{Accuracy Paradox.~\cite{Zemblys2018GazeNetNetworks}} Event matching techniques based on the largest overlap ratio, such as the event $\kappa$ proposed by Zemlys \etal do not  provide a reliable measure of alignment quality.~\cite{Zemblys2018GazeNetNetworks} ELC overcomes these issues by matching events whose transition points fall within a window. A potential drawback of ELC is its dependency on the window size. Although the window size could be carefully chosen for different types of events and transitions, the metric could generate different results due to varying window sizes. For example, if a small window size was chosen, ELC would have a lower tolerance for transition ambiguity between certain event types which could result in higher misclassification scores. Furthermore, ELC is not symmetrical. To alleviate that, we propose that metrics derived using ELC should be averaged when used to evaluate inter-coder performance. While ELC overcomes certain drawbacks from previous evaluation techniques, new event level metrics are needed which accurately reflect performance, is symmetric in nature, provides a reliable measure of temporal alignment quality and is independent of an external threshold.

\section{Conclusion}
This work introduces GW, a large-scale dataset for studying eye and head coordination in naturalistic conditions. Participants were asked to perform four tasks without constraining them in any manner and were free to accomplish the tasks in any manner they chose to. Approximately 2 hours and 15 minutes of gaze behavior was manually hand coded by multiple human annotators and used to train gaze classifiers. We benchmark the performance of two machine learning algorithms for classifying these events and found that both achieved near human level performance for detecting gaze fixations and saccades, but they found it difficult to distinguish gaze pursuit behavior without additional contextual information otherwise available to human coders. In an effort to produce intuitive measures for event level similarities between two sequences, we propose the ELC event matching algorithm. We verify that all commercial eye tracking solutions could benefit in classifying head-free gaze pursuit movements by including a low cost IMU. Furthermore, comparable results are observed when head-free gaze movements are classified purely based on absolute velocity information of the eye and head, which indicates that head-free gaze classification is possible without aligning the eye and head coordinate systems.

\label{stats}



\section{Acknowledgements}
The authors would like to thank the Google Daydream team for funding this research.

\section{Author contributions}
All authors contributed to drafting the paper, data interpretation, project development, and approved the final version of the manuscript for submission. R.S.K. created the hardware setup, data collection, and analysis. Z.Y developed the evaluation metrics.

\section{Additional information}
\textbf{Competing interests:} The authors declare no competing interests.\\
\textbf{Dataset availability:} Compressed data and codes will be made publicly available.\\
\textbf{Permission to share facial imagery:} All identifiable people in this manuscript consent to share their information as presented.
\textbf{Data representation:} All statistical figures were generated using Gramm, an open source software for data visualization.~\cite{morel2018gramm} All metrics reported are rounded to the second decimal.
\section{Supplementary Information}
\textbf{Manuscript Title}: Gaze-in-wild: A dataset for studying eye and head coordination in everyday activities\\
\textbf{Author list}: Rakshit Kothari, Zhizhuo Yang, Christopher Kanan, Reynold Bailey, Jeff Pelz, Gabriel Diaz
\setcounter{figure}{0}
\setcounter{table}{0}

\begin{table}
\centering
\renewcommand{\arraystretch}{1.5}
{%
\begin{tabular}{cl|c|c|c|c|c|c|c|c|c|c|c|c}
\hline
Person & Age & \multicolumn{3}{c|}{Indoor navigation}                                        & \multicolumn{3}{c|}{Ball catching}                                            & \multicolumn{3}{c|}{Visual search}                           & \multicolumn{3}{c|}{Tea making}                                    \\ \hline
       &     & Status                      & $\Delta^{ETG}_\theta$ & $\Delta^{IMU}_{\theta}$ & Status                      & $\Delta^{ETG}_\theta$ & $\Delta^{IMU}_{\theta}$ & Status     & $\Delta^{ETG}_\theta$ & $\Delta^{IMU}_{\theta}$ & Status           & $\Delta^{ETG}_\theta$ & $\Delta^{IMU}_{\theta}$ \\ \hline
1      & 27  & $\doublecheck$              & 0.75                  & 8.71                    & $\doublecheck$              & 0.67                  & 7.90                     & $\times$   &                       &                         & $\times$         &                       &                         \\
2      & 18  & $\overset{+}{\doublecheck}$ & 0.79                  & 0.76                    & $\overset{+}{\doublecheck}$ & 0.89                  & 6.63                    & $\times$   &                       &                     & $\checked$   & 0.76                  &                         \\
3      & 19  & $\checked$                  & 1.27                  & 0.05                    & $\overset{+}{\checked}$            & 0.75                  & 2.41                    & $\times$   &                       &                         & $\times$         &                       &                         \\
6      & 18  & $\checked$                  & 0.43                  & 45.01$^\gamma$          & $\doublecheck$              & 0.37                  & 10.11                   & $\times$   &                       &                     & $\overset{+}{O}$ & 0.69                  & 0.84                     \\
8      & 22  & $\checked$                  & 0.99                  & 4.67                    & $\checked$                  & 1.29                  & 1.91                    & $\checked$ & 0.98                  & 0.56                    & $\times$         &                       &                         \\
9      & 22  & $\checked$                  & 0.62                  & 3.66$^\gamma$           & $\checked$                         & 0.73                  & 2.95                    & $O$        & 0.59                  & 0.27                    & $\times$         &                       &                         \\
10     & 26  & $\checked$                  & 0.55                  & 13.00                      & $O$                         & 1.29                  & 7.36                    & $O$        & 1.04                  & 0.01                    & $O$              & 0.49                  & 0.73                    \\
11     & 25  & $O$                         & 0.77                  & 31.96$^\gamma$          & $O$                         & 0.56                  & 4.03                    & $\times$   &                       &                         & $O$              & 0.69                  & 4.63                    \\
12     & 26  & $\checked$                  & 0.52                  & 2.83                    & $\checked$                  & 0.65                  & 2.04                    & $\checked$ & 0.75                  & 1.17                    & $\checked$       & 0.57                  & 12.17                   \\
13     & 27  & $O$                         & 1.14                  & 3.70                     & $O$                         & 1.08                  & 0.44                    & $O$        & 0.83                  & 7.52                    & $O$              & 0.80                  & 14.14                   \\
14     & 23  & $O$                         & 0.88                  & 0.34                    & $O$                         & 1.20                  & 0.02                    & $O$        & 0.82                  & 1.63                    & $O$              & 1.08                  & 1.19                    \\
15     & 22  & $O$                         & 1.09                  & 8.33                    & $O$                         & 0.90                  & 1.33                    & $\checked$ & 0.73                  & 15.56$^\gamma$                   & $O$              & 0.67                  & 7.43                    \\
16     & 22  & $\checked$                  & 0.82                  & 0.41                    & $\checked$                         & 0.55                  & 0.53                    & $O$        & 0.94                  & 5.31                    & $O$              & 0.92                  & 0.88                    \\
17     & 23  & $\checked$                  & 1.01                  & 4.28                    & $\checked$                         & 0.89                  & 0.22                    & $\checked$ & 0.65                  & 0.22                    & $O$              & 0.79                  & 22.76$^\gamma$          \\
18     & 34  & $\checked$                         & 1.01                  & 14.46                   & $O$                         & 0.99                  & 1.63                    & $\checked$        & 0.71                  & 12.09                   & $\checked$              & 0.81                  & 3.07                    \\
19     & 26  & $O$                  & 0.88                  & 2.38                    & $\checked$                  & 0.73                  & 6.52                    & $\checked$ & 0.96                  & 19.34$^\gamma$                   & $\times$         &                       &                         \\
20     & 55  & $\checked$                  & 0.93                  & 7.01                    & $O$                         & 1.23                  & 3.58                    & $\checked$ & 0.95                  & 20.47$^\gamma$                   & $O$       & 1.24                  & 2.87                    \\
22     & 54  & $\checked$                  & 0.50                  & 6.84                    & $\checked$                  & 0.38                  & 6.89                    & $\checked$ & 0.55                  & 11.61                   & $\times$       &                   &                    \\
23     & 60  & $O$                         & 1.23                  & 7.45                    & $O$                         & 0.47                  &    3.50                     & $O$        & 0.44                  & 2.04                    & $O$              & 0.44                  & 2.54                   
\end{tabular}
}
\caption{Dataset status and error measures. $\Delta^{ETG}_{\Theta}$ $\rightarrow$ Calibration error in degrees $^\circ$. $\Delta^{IMU}_{\Theta}$ $\rightarrow$ IMU absolute angular drift measured by the difference in the head orientation at the start and end of a task in $^\circ$. Note that certain recordings do not have this measure because we did not explicitly constrain the participants to arrive at their initial pose. For such situations, we estimate head pose when subjects were approximately fixated towards an identifiable direction using visual imagery in the middle of a task. Status symbols: $\doublecheck$ $\rightarrow$ recording has labels from multiple labellers. $\checked$ $\rightarrow$ recording is labeled by a single person. $O$ $\rightarrow$ Not labeled. $+$ $\rightarrow$ Depth values not present. $\gamma$ $\rightarrow$ Manual head pose correction. $\times$ $\rightarrow$ Data discarded.}
\label{tbl:Dataset_status}
\end{table}

\begin{figure}
\centering
\includegraphics[width=\linewidth]{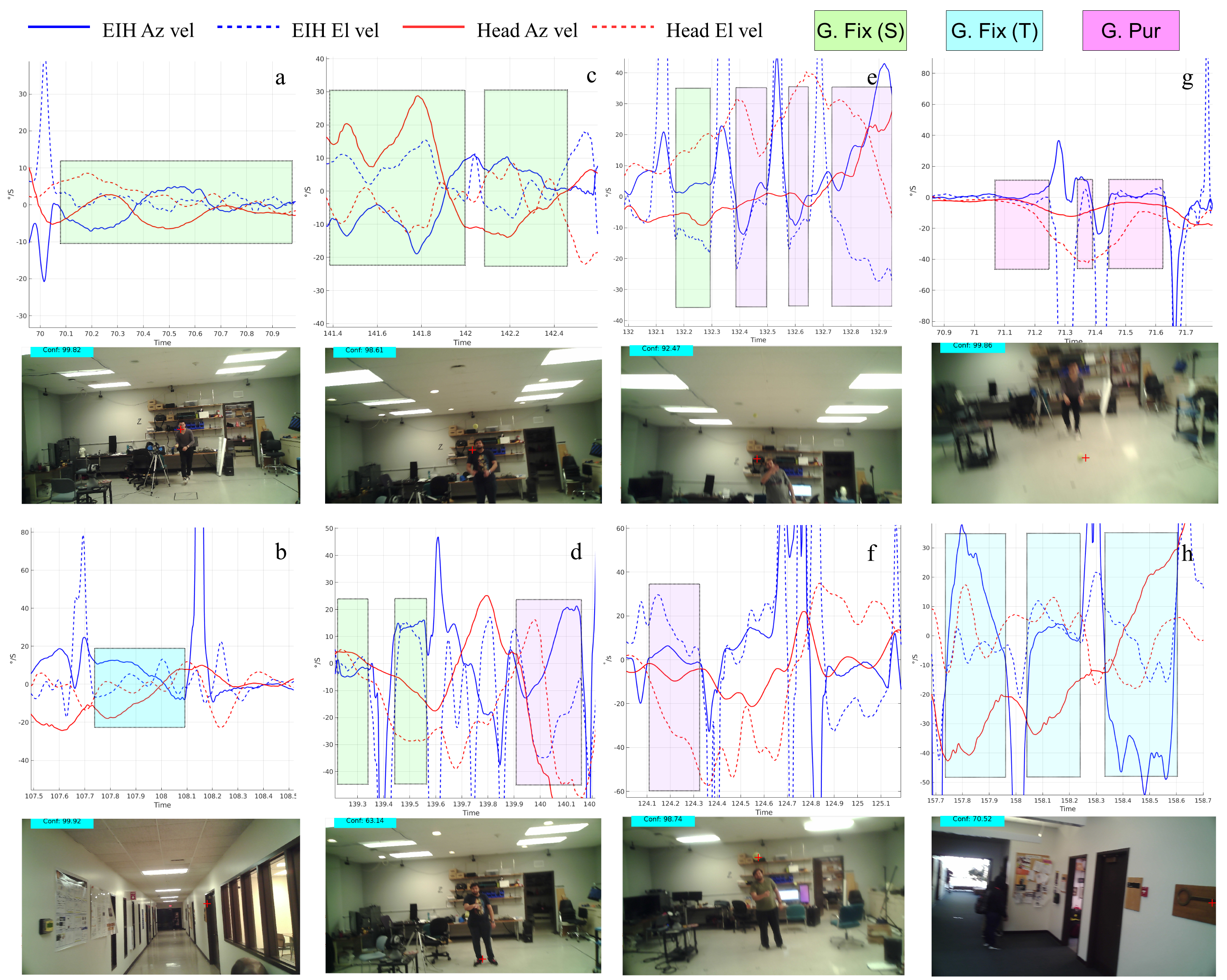}
\caption{Examples of gaze behavior in 1$s$ clips. Eye and head velocity signals and the related world camera view which best describes the signals is provided as a pair in each row. The bold blue line signifies EiH azimuthal velocity while the dashed line signifies EiH elevation velocity. Likewise, the red variant stands for head velocity traces. Rectangular boxes within each sub-figure are used to draw the readers attention to marked events. Sub-figure \textbf{a} and \textbf{b} signify gaze fixation events. Note that due to a slight misalignment in the IMU and eye tracker, the velocity components in \textbf{a} do not precisely cancel out. Sub-figure \textbf{c} and \textbf{d} showcase gaze fixation events while the head catches up to the fixation location. Notice that in \textbf{d}, the EiH and head velocity traces do not cancel out despite being marked as a fixation. Signal information alone convey that the traces are a head dominated pursuit movement, however, the world imagery shows that the head is catching up to the fixation location. Sub-figures \textbf{e}, \textbf{f} and \textbf{g} highlight examples of head dominated pursuit behavior punctuated by saccades. Note the fixation event marked in \textbf{e}. It appears that the eye and head signals compensate in the elevation component but head movements dominate in the azimuthal direction. In hindsight, it appears to be part of a sequence of pursuit events interjected by predictive saccades but has been incorrectly marked as a fixation. Sub-figure \textbf{h} is an interesting example of fixation under translation. Visual imagery conveys that the underlying behavior is fixational but one can see that the velocity traces resembles that of a pursuit movement. Examples listed in \textbf{d} and \textbf{h} contribute heavily to fixation/pursuit confusion.}
\label{fig:example_signals}
\end{figure}

\begin{table}
\centering
\renewcommand{\arraystretch}{1.5}
\begin{tabular}{|c|c|c|c|c|c|c|c|}
\hline
 & Metric & biRNN-1 & biRNN-2 & biRNN-3 & biRNN-4 & \begin{tabular}[c]{@{}c@{}}biRNN-3\\  (only eyes)\end{tabular} & \begin{tabular}[c]{@{}c@{}}biRNN-3\\  (only absolute)\end{tabular} \\ \hline
\multirow{4}{*}{\begin{tabular}[c]{@{}c@{}}Sample\\ level\\ metrics\end{tabular}} & Overall $\kappa$ & 0.62 & 0.64 & 0.65 & 0.65 & 0.62 & 0.64 \\
 & G. Fix $\kappa$ & 0.61 & 0.63 & \textbf{0.65} & 0.64 & 0.61 & 0.63 \\
 & G. Pur $\kappa$ & 0.34 & 0.37 & \textbf{0.4} & 0.36 & 0.29 & 0.38 \\
 & Sac $\kappa$ & 0.69 & 0.7 & 0.7 & 0.69 & 0.7 & 0.7 \\ \hline
\multirow{10}{*}{\begin{tabular}[c]{@{}c@{}}Event\\ level\\ metrics\end{tabular}} & Overall ($\kappa$, $\kappa$*) & 0.51, 0.43* & 0.53, 0.46* & \textbf{0.54, 0.47*} & 0.52, 0.46* & 0.52, 0.46* & 0.54, 0.46* \\
 & G. Fix ($\kappa$, $\kappa$*) & 0.65, 0.4* & 0.66, 0.43* & \textbf{0.67, 0. 45*} & 0.66, 0.44* & 0.65, 0.43* & 0.67, 0.43* \\
 & G. Pur ($\kappa$, $\kappa$*) & 0.23, 0.17* & 0.28, 0.20* & 0.30, 0.24* & 0.30, 0.26* & 0.14, 0.17* & 0.24, 0.21* \\
 & Sac ($\kappa$, $\kappa$*) & 0.7, 0.5* & 0.72, 0.53* & 0.72, 0.53* & 0.70, 0.51* & 0.72, 0.54* & 0.73, 0.53* \\
 & G. Fix F$_1$ & 0.83 & 0.83 & \textbf{0.84} & 0.83 & 0.83 & 0.83 \\
 & G. Pur F$_1$ & 0.35 & 0.39 & \textbf{0.42} & 0.37 & 0.3 & 0.39 \\
 & Sac F$_1$ & 0.85 & 0.86 & 0.86 & 0.85 & 0.86 & 0.86 \\
 & G. Fix $l_2$ ($\mu$, $\sigma$) & 15.62, 4.00 & 15.14, 3.67 & \textbf{15.11, 3.47} & 15.32, 3.58 & 15.18, 3.53 & 15.13, 3.56 \\
 & G. Pur $l_2$ ($\mu$, $\sigma$) & 15.83, 4.08 & 14.77, 4.41 & \textbf{12.97, 4.17} & 13.27, 3.70 & 13.06, 4.44 & 14.06, 3.87 \\
 & Sac $l_2$ ($\mu$, $\sigma$) & 15.52, 4.11 & 14.89, 3.66 & 14.86, 3.49 & 14.97, 3.66 & 14.86, 3.57 & \textbf{14.83, 3.60} \\
 & G. Fix $O_r$ & 0.92 & 0.92 & 0.92 & 0.92 & 0.92 & 0.92 \\
 & G. Pur $O_r$ & 0.89 & 0.91 & 0.91 & 0.91 & 0.92 & 0.92 \\
 & Sac $O_r$ & 0.69 & 0.70 & 0.70 & 0.70 & 0.70 & 0.70 \\ \hline
\end{tabular}
\caption{Ablation results reported using all metrics. Note that event level $\kappa$* scores are computed using the ELC matching technique as described in Section \ref{event_metric}. Mean $l_2$ distance and it's deviation, $l_2-\sigma$ between between matched events are reported in $ms$.}
\label{tbl:ablation_full}
\end{table}

\end{document}